%% file: main.tex
\documentclass{style/naturep}

\usepackage{amssymb}
\usepackage{amsmath}
\usepackage{graphicx}
\usepackage{algorithmicx}
\usepackage{algorithm}
\usepackage{adjustbox}
\usepackage{graphicx}
\usepackage{float}
\usepackage{hyperref}
\usepackage{xurl}

\usepackage[noend]{algpseudocode}
\usepackage{bbm}
\usepackage{lineno}
\usepackage[margin=0.6in]{geometry}
\usepackage{ragged2e}
\usepackage{setspace}
\usepackage{longtable}
\usepackage{hyperref}
\usepackage{anyfontsize}
\usepackage{setspace}
\usepackage{float}
\usepackage{booktabs}
\usepackage{multirow}
\usepackage{blindtext}
\usepackage{tabularx}
\usepackage{caption}
\usepackage{subcaption}
\usepackage{enumitem, kantlipsum}
\usepackage{xspace}
\usepackage{pifont}

\usepackage{graphicx}
\usepackage{amsmath}
\usepackage{amssymb}
\usepackage{booktabs}
\usepackage{float}
\usepackage{placeins}
\usepackage[table,x11names]{xcolor}
\definecolor{maroon}{cmyk}{0,0.87,0.68,0.32}
\definecolor{gray}{rgb}{0.3,0.3,0.3}

\newcommand\Heading[1]{
  \noindent\textbf{\Large{#1}}
}

\newcommand\heading[1]{
  \noindent\textbf{\large{#1}}
}

\title{\begin{flushleft}{\begin{spacing}{1}
   Unified integration of pathology foundation models for scalable histopathology analysis
\end{spacing}}\end{flushleft}}

\makeatletter
\let\saved@includegraphics\includegraphics
\AtBeginDocument{\let\includegraphics\saved@includegraphics}

\makeatother

\begin{document}

\input{sections/0-cover_page}
\clearpage


\begin{spacing}{1.35}
\Heading{Introduction}
\input{sections/1-introduction}

\Heading{Results}

\input{sections/2-performance}

\Heading{Discussion}

\input{sections/3-discussion}

\end{spacing}

\begin{spacing}{1.35}
\Heading{Online Methods}
\input{supplement/1-framework}

\input{supplement/2-datasets}

\input{supplement/3-additional}

\end{spacing}

\clearpage

\setcounter{table}{0}
\renewcommand{\tablename}{Extended Data Table}

\clearpage
\begin{nolinenumbers}
\Heading{References}

\vspace{2mm}

\begin{spacing}{0.9}
\bibliographystyle{naturemag}
\bibliography{main}
\end{spacing}
\end{nolinenumbers}
\clearpage

\end{document}

%% file: sections/0-cover_page.tex
\maketitle
\vspace{-20mm}
\begin{spacing}{1.4}
\noindent Wenhui Lei$^{1\boldsymbol{\ddag}}$, Yusheng Tan$^{2\boldsymbol{\ddag}}$, Anqi Li$^{2\boldsymbol{\ddag}}$, Hengrui Tian$^{1}$, Ruiying Li$^{3}$, Zhenqun Jiang$^{1}$, Fang Yan$^{4}$, Hanyu Chen$^{*5}$, Xiaofan Zhang$^{*1,6}$, Shaoting Zhang$^{*1,7}$
\end{spacing}

\vspace{-7mm}
\begin{spacing}{1.4}
\begin{affiliations}
 \item Shanghai Jiao Tong University, Shanghai, China
 \item Washington University in St. Louis, St. Louis, USA
 \item University of Science and Technology Beijing, Beijing, China
 \item Shanghai Artificial Intelligence Laboratory, Shanghai, China
 \item Department of Surgical Oncology and General Surgery, Key Laboratory of Precision Diagnosis and
Treatment of Gastrointestinal Tumours, Ministry of Education, The First Hospital of China Medical University, Liaoning, China
 \item Shanghai Innovation Institute, Shanghai, China
 \item Sensetime Research, Shanghai, China
 \\$\boldsymbol{\ddag}$ Contributed Equally
 \\\textbf{*Corresponding author}: Hanyu Chen (hychen88@cmu.edu.cn), Xiaofan Zhang (xiaofan.zhang@sjtu.edu.cn), Shaoting Zhang (shaoting.zhang@sjtu.edu.cn)
\end{affiliations}
\end{spacing}

\vspace{-3mm}
\begin{spacing}{1.2}
\noindent \textbf{Foundation models have substantially advanced computational pathology by learning transferable visual representations from large histological datasets, yet their performance varies widely across tasks due to differences in training data composition and reliance on proprietary datasets that cannot be cumulatively expanded. Existing efforts to combine foundation models through offline distillation partially mitigate this issue but require dedicated distillation data and repeated retraining to integrate new models. Here we present Shazam, an online integration model that adaptively combines multiple pretrained pathology foundation models within a unified and scalable representation learning paradigm. Our findings show that fusing multi-level features through adaptive expert weighting and online distillation enables efficient consolidation of complementary model strengths without additional pretraining. Across spatial transcriptomics prediction, survival prognosis, tile-level classification, and visual question answering, Shazam consistently outperforms strong individual models, demonstrating that online model integration provides a practical and extensible strategy for advancing computational pathology.
}
\end{spacing}

%% file: sections/1-introduction.tex
\paragraph{} Histopathology remains the cornerstone of cancer diagnosis and treatment planning, providing direct visualization of tissue architecture, cellular morphology, and tumor–microenvironment interactions that guide clinical decision-making. Advances in whole-slide imaging (WSI) have transformed this traditionally microscope-based discipline into a scalable digital workflow, enabling the systematic capture, storage, and analysis of gigapixel-resolution slides. This digital transition has laid the foundation for modern computational pathology (CPath), where artificial intelligence systems can interrogate WSI data to support diagnostic, predictive, and prognostic tasks\cite{van2021deep,shmatko2022artificial}. Building on this momentum, recent pathology foundation models have introduced a new paradigm for CPath. By pretraining on millions of tiles extracted from large cohorts of WSIs across diverse organs and cancer types, models such as Virchow\cite{vorontsov2024foundation}, UNI\cite{chen2024towards}, and H-Optimus-1\cite{hoptimus1} learn broad and transferable representations that enhance data efficiency and improve downstream performance across a wide spectrum of clinical tasks.

\paragraph{} Despite these advances, recent benchmarking studies have revealed substantial performance variability among pathology foundation models across tasks\cite{campanella2025clinical,neidlinger2025benchmarking,ma2025pathbench}. Much of this inconsistency reflects differences in the scale, diversity, and institutional origins of the pretraining data, and no single model consistently performs well across diagnostic, prognostic, molecular, and region-level tasks. Compounding this challenge, many high-performing models rely on large proprietary datasets that cannot be shared due to privacy and institutional constraints\cite{xu2024whole,chen2024towards,hoptimus1}. This lack of data accessibility prevents the research community from steadily expanding the pretraining corpus over time, constraining the growth in data diversity that has supported advances in natural image and language foundation models\cite{deng2009imagenet,schuhmann2022laion,commoncrawl_aws}.

One line of recent work has sought to mitigate limited access to large-scale pathology training data by combining existing foundation models through offline knowledge distillation, exemplified by GPFM\cite{ma2025generalizable}. In this framework, multiple pretrained teacher models first generate features on a curated distillation dataset, and a student model is trained to learn from these signals through expert distillation, complemented by self-distillation that aligns local and global representations. While this strategy provides a practical means of integrating knowledge from several models, it faces several inherent limitations. First, because the distillation process is performed offline, incorporating newly developed foundation models requires retraining the student model from scratch, making rapid integration of emerging advances difficult. Second, the effectiveness of offline distillation depends heavily on the scale and diversity of the available distillation dataset, which in practice is restricted to publicly accessible or newly assembled collections. This constraint limits the range of morphologic and clinical variation that the student model can encounter during distillation, reducing its ability to fully capture the complementary behaviors of diverse teacher models\cite{fang2026knowledge}. Third, offline distillation provides limited flexibility for task-specific adaptation, since the student model learns from teacher outputs generated in a task-agnostic manner and cannot selectively emphasize teacher knowledge that is particularly relevant for a specific downstream prediction target. These constraints highlight the need for an alternative strategy that can directly leverage multiple foundation models in a more flexible and adaptive way, without relying on repeated offline retraining or large-scale distillation corpora.

\paragraph{} To directly address these challenges, we developed Shazam, a unified ensemble model that integrates multiple pathology foundation models within a single flexible inference system (Fig.~\ref{fig:fig1}a). In contrast to offline distillation approaches that require assembling a dedicated distillation dataset and retraining a new model, Shazam performs task-specific online integration. It combines multi-level representations from diverse pathology foundation models through a lightweight Mixture-of-Experts (MoE)\cite{fedus2022switch} mechanism and adapts the fusion process directly to the target task. This design enables Shazam to extract the most informative elements from each model without the need to restart pretraining or construct large-scale distillation corpora whenever new foundation models become available. We evaluate Shazam across a wide range of downstream applications, including spatial transcriptomics prediction, whole-slide survival prognosis, tile-level classification, and visual question answering. These tasks span molecular, regional, and slide-level endpoints and together reflect the breadth of clinical challenges encountered in computational pathology. Across these evaluations, Shazam demonstrates consistent and robust performance. In total, it achieves an average ranking score of 1.17 across 30 benchmarks, substantially outperforming the second-best model, Virchow2\cite{zimmermann2024virchow2}, which attains an average of 3.20. By enabling task-adaptive, online integration of heterogeneous pretrained models, Shazam provides a practical mechanism for consolidating complementary representations without requiring additional large-scale pretraining or privileged data access. More broadly, this paradigm suggests a shift in pathology foundation modeling from isolated, monolithic models toward modular and compositional systems, offering a scalable path for continual methodological evolution under realistic clinical and institutional constraints.

\includegraphics[width=\linewidth]{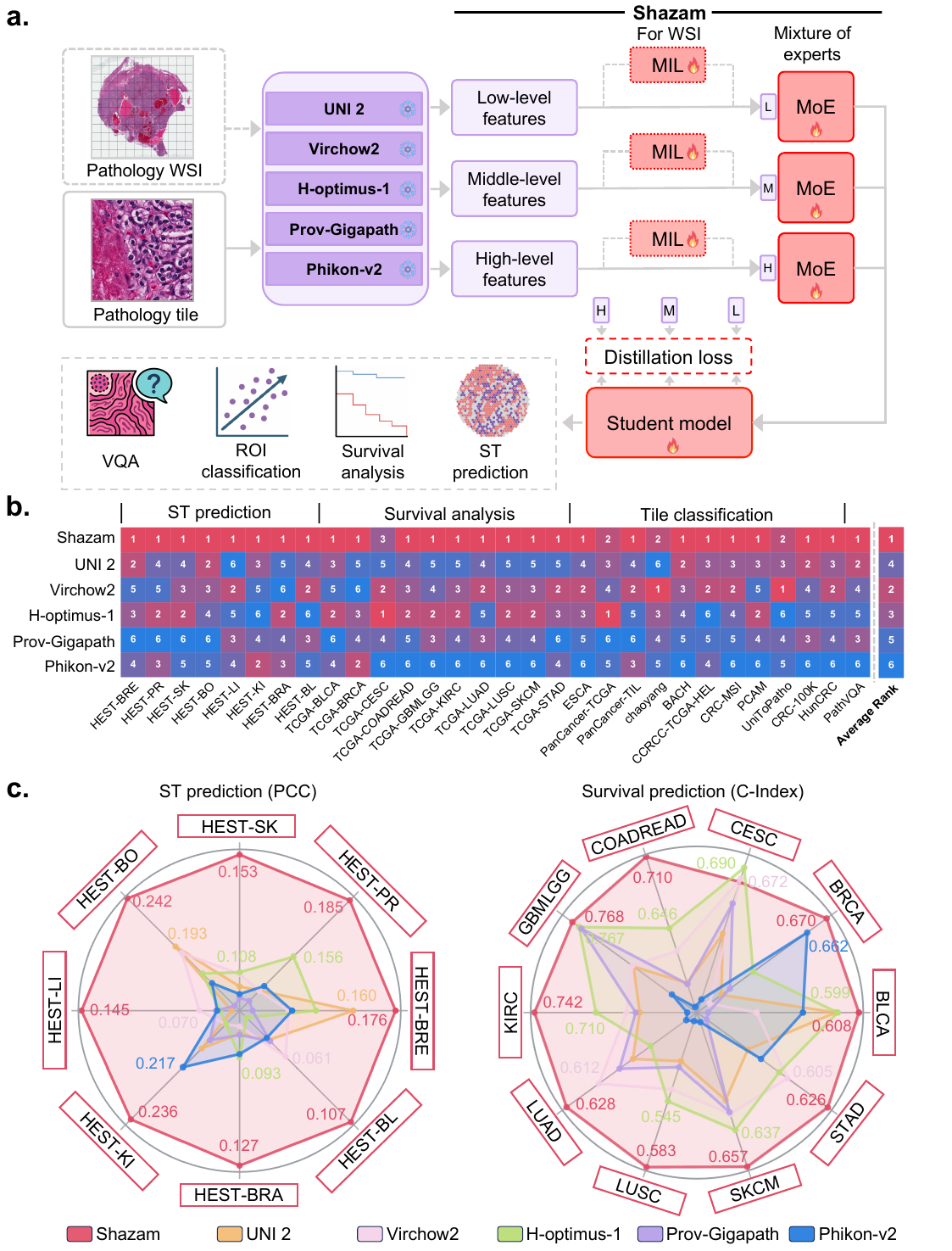}
\captionof{figure}{\textbf{a}, Overview of the Shazam framework. Multiple pretrained pathology foundation models are used to extract features from WSIs and tiles. For WSI-level tasks, each model first produces slide representations through a multiple-instance-learning module, whereas tile-level tasks directly use tile embeddings. Shazam performs task-specific online feature integration by fusing low-, middle-, and high-level features and assigning adaptive mixture-of-experts weights. A student model is optimized with an online distillation objective to learn cohesive task-aligned representations without the need for offline distillation or retraining. \textbf{b}, Overall ranking of Shazam and baseline foundation models across 30 downstream benchmarks. A lower rank indicates stronger performance, with a rank of 1 assigned to the top-performing model for each task. 
\textbf{c}, Comparison of model performance in spatial transcriptomics and survival prediction measured by PCC and C-index. \\[1em]}
\label{fig:fig1}

%% file: sections/2-performance.tex
Shazam is built upon five state-of-the-art pathology foundation models, including UNI 2\cite{chen2024towards}, Virchow2\cite{zimmermann2024virchow2}, H-optimus-1\cite{hoptimus1}, Prov-Gigapath\cite{xu2024whole}, and Phikon-v2\cite{filiot2024phikonv2largepublicfeature}, and is evaluated across a broad spectrum of downstream applications. Our benchmark covers four major task categories: spatial transcriptomics prediction with eight tasks, whole-slide survival analysis with ten tasks, tile-level classification with eleven tasks, and pathology visual question answering. The overall ranking across all thirty tasks is summarized in Fig. \ref{fig:fig1}b. Shazam achieves the strongest average ranking score of 1.17 and ranks first on 26 of the 30 tasks. By comparison, the next best model, Virchow2, attains an average ranking score of 3.20 and leads only 2 tasks. These findings demonstrate that Shazam substantially outperforms the foundation models on which it is built and exhibits strong generalization across diverse molecular, regional, and slide-level prediction settings. 


\section{Shazam enhances spatial transcriptomic prediction}

\paragraph{}Spatial transcriptomics (ST) provides high-resolution maps of gene expression that reveal the molecular architecture of tissues and the tumor microenvironment, offering insights beyond what morphology alone can capture\cite{staahl2016visualization,baccin2020combined,janesick2023high}. However, the high cost of ST assays and the limited availability of standardized paired H\&E--ST datasets have traditionally restricted most computational analyses to narrow tasks and small patient cohorts\cite{meylan2022tertiary,bassiouni2023spatial,parigi2022spatial}. Leveraging pathology foundation models to infer spatial gene expression directly from H\&E images offers a scalable alternative, enabling the discovery of treatment-relevant biomarkers across much larger datasets. To assess performance on this challenging task, we compared Shazam against a suite of strong pathology foundation-model baselines for predicting spatial gene expression from H\&E slides in the HEST-1k dataset\cite{jaume2024hest}, focusing on clinically actionable genes as prediction targets. We considered eight organ-specific cohorts with sufficient paired data, including bowel, liver, breast, prostate, skin, brain, kidney, and bladder, and curated clinically actionable gene sets for each cohort. Additional details are provided in Methods and Supplementary Table\ref{tab:cancer_gene_lists_csv_1}-\ref{tab:cancer_gene_lists_csv_2}.

\paragraph{}For each cohort, we followed the patient-level splitting strategy and conducted patient-level \(K\)-fold cross-validation, formulating gene expression inference as a tile-level regression task (details in Methods). For each baseline foundation model, we attached a multilayer perceptron to its pretrained representations to predict log-normalized spot-level gene expression. Shazam adopted the same regression setting but first integrated representations from multiple foundation models into a unified feature space before prediction with the multilayer perceptron. We quantified performance using the Pearson Correlation Coefficient (PCC) between predicted and measured spot-level expression profiles, consistent with the evaluation protocol of HEST-1k as described in Methods.

\paragraph{} Across all eight organ types, Shazam consistently achieved the highest PCCs, with an average improvement of 0.08–0.17 over the strongest individual foundation models (Fig.~\ref{fig:fig1}a). The baseline models exhibit clear organ-dependent variability, with Virchow2 performing relatively well in brain and skin cancers, H-optimus-1 showing stronger results in prostate and bladder cancers, and UNI 2 occasionally ranking higher in bowel or kidney cancers (details in Supplementary Tables \ref{tab:Table2},\ref{tab:Table3}). These differences indicate that existing pathology foundation models often reflect the tissue compositions and morphological patterns emphasized during pretraining, which makes it challenging for a single model to maintain uniformly strong performance across distinct organs. In contrast, Shazam ranks first in all eight cancer types, achieving an average ranking of 1.0 and significantly outperforming the next-best models, including UNI 2, Virchow2, and H-optimus-1, whose mean ranking scores are approximately 3.25 ($p$=0.004). This consistent advantage across glandular, stromal-rich, and neuroepithelial tissues reflects Shazam’s ability to integrate complementary representations from multiple expert models and to provide stable, cross-organ generalization. In addition to cohort-level performance, Fig.~\ref{fig:fig2}b presents a gene-wise comparison of PCC values for spatial expression prediction in the HEST bowel cancer cohort. Shazam achieves consistently higher correlations than all baseline models across the majority of target genes, indicating that its performance gains are not driven by a small subset of genes but instead reflect a stable improvement across diverse molecular targets. Similar gene-wise PCC trends are observed in other cancer types and are provided in Supplementary Figs. \ref{fig:hest-gene-1}-\ref{fig:hest-gene-5}.

\paragraph{} Consistent with these quantitative results, Fig.~\ref{fig:fig2}c shows a representative bowel cancer case from the HEST-1k dataset. Shazam accurately captured the heterogeneous spatial pattern of IDH1 expression, including the pronounced up-regulation in dysplastic tumor glands relative to adjacent normal glandular epithelium. This tumor-specific elevation is clearly reflected in the ground truth and was faithfully reproduced only by Shazam, whereas baseline models such as UNI 2, Virchow2, H-optimus-1, GigaPath, and Phikon-v2 either failed to distinguish these compartments or produced the opposite trend. Shazam also recovered the expected low-expression domains within stromal and muscular layers, which other models tended to oversmooth or underrepresent. These results illustrate Shazam’s ability to align tissue morphology with gene-level variation, capturing both tumor-associated up-regulation and stromal or muscular down-regulation patterns that single-model architectures frequently mislocalize.

\paragraph{} Collectively, these results demonstrate that Shazam effectively learns a unified morphomolecular representation, translating histological features into gene-level spatial expression patterns with superior accuracy and 

\includegraphics[width=1.0\linewidth]{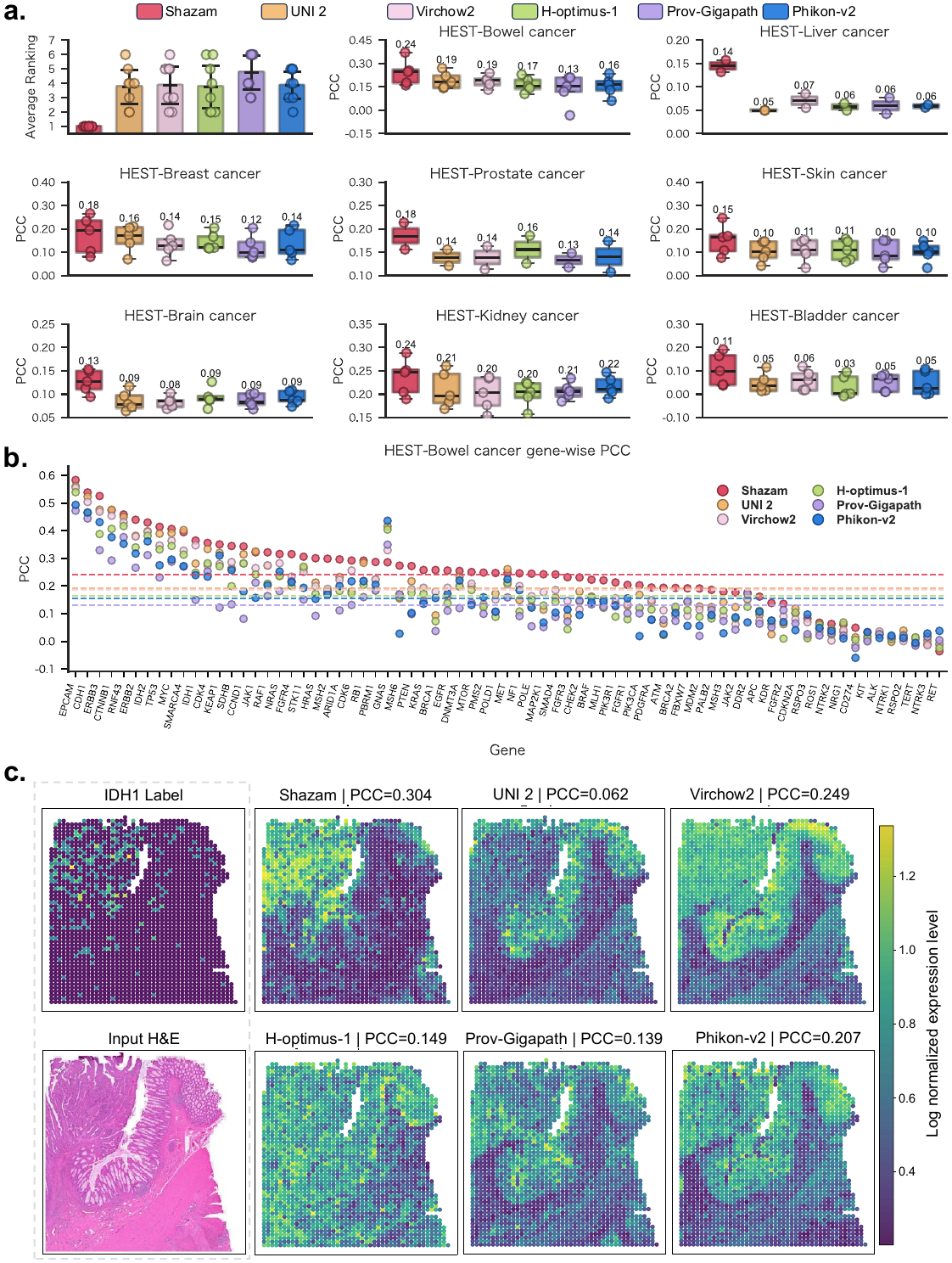}
\captionof{figure}{
\textbf{Evaluation of spatial molecular expression prediction across organs and genes.}
\textbf{a}, Quantitative comparison of prediction accuracy across eight organ types in the HEST-1k dataset, measured by Pearson Correlation Coefficient (PCC). Shazam consistently outperforms existing pathology foundation models (UNI 2, Virchow2, H-optimus-1, Phikon-v2, GigaPath). Box plots show the median (center line), interquartile range (IQR, box limits represent 25th and 75th percentiles), and whiskers extending to 1.5 × IQR; outliers are shown as individual points.
\textbf{b}, Gene-wise PCC for spatial expression prediction in the HEST-Bowel cancer cohort, comparing Shazam with baseline pathology foundation models.
\textbf{c}, Spatial prediction of IDH1 expression in a colorectal cancer sample. Shazam accurately recapitulates the elevated expression within dysplastic tumor glands and the reduced expression in stromal and muscular layers, aligning closely with the measured spatial transcriptomic profile. Baseline models fail to recover these tissue-compartment differences.}
\label{fig:fig2}

\vspace{1.0em}
cross-tissue generalization. Taken together, these findings show that Shazam provides a practical and reliable framework for spatial transcriptomic prediction, offering a flexible foundation for downstream molecular analysis without overreliance on costly assays.

\section{Shazam advances survival prognosis}
\paragraph{}Accurate survival prognosis is central to cancer management, guiding treatment selection, surveillance strategies, and clinical trial allocation\cite{amin2017eighth}. Yet substantial variability in patient outcomes persists even within the same cancer type, underscoring the limitations of existing clinical and genomic biomarkers. Histopathology, obtained routinely at diagnosis and rich in biological information, remains an underleveraged resource for prognostic modeling. To assess whether Shazam’s unified morphomolecular representations can capture outcome-relevant signals, we conducted a pan-cancer evaluation across ten The Cancer Genome Atlas Program (TCGA)\cite{weinstein2013cancer} cohorts. For each cohort, diagnostic whole-slide images were processed to obtain slide-level representations using attention-based multiple instance learning (details in Methods), and Cox proportional hazards models were trained in a five-fold patient-level cross-validation framework to evaluate the generalizability of survival predictions derived directly from H\&E slides.

\paragraph{}Across the ten cancer types, Shazam achieved the strongest overall prognostic performance, attaining a mean C-index of 0.666 (Supplementary Tables \ref{tab:Table5}-\ref{tab:Table7}). This surpasses the best-performing individual foundation model, H-optimus-1 (0.642), and exceeds the average performance of Virchow2 (0.631), Prov-GigaPath (0.617), UNI 2 (0.617), and Phikon-v2 (0.594). A paired Wilcoxon signed-rank test confirmed that Shazam’s improvement over H-optimus-1 is statistically significant ($p = 0.007$), indicating a consistent advantage rather than cohort-specific variability. Shazam delivered noticeable gains in several cancer types with heterogeneous histology, including colon adenocarcinoma (COAD), where the C-index increased from 0.646 (H-optimus-1) to 0.710, and lung squamous cell carcinoma (LUSC), where it rose from 0.545 (H-optimus-1) to 0.583. These improvements highlight Shazam’s capacity to integrate complementary strengths from multiple foundation models, yielding a more comprehensive and discriminative representation of disease progression than any individual model.

\paragraph{}To further evaluate the clinical relevance of these quantitative gains, we performed Kaplan–Meier analyses using Shazam-derived risk scores. Patients were stratified into high-risk and low-risk groups using a median 

\includegraphics[width=1.0\linewidth]{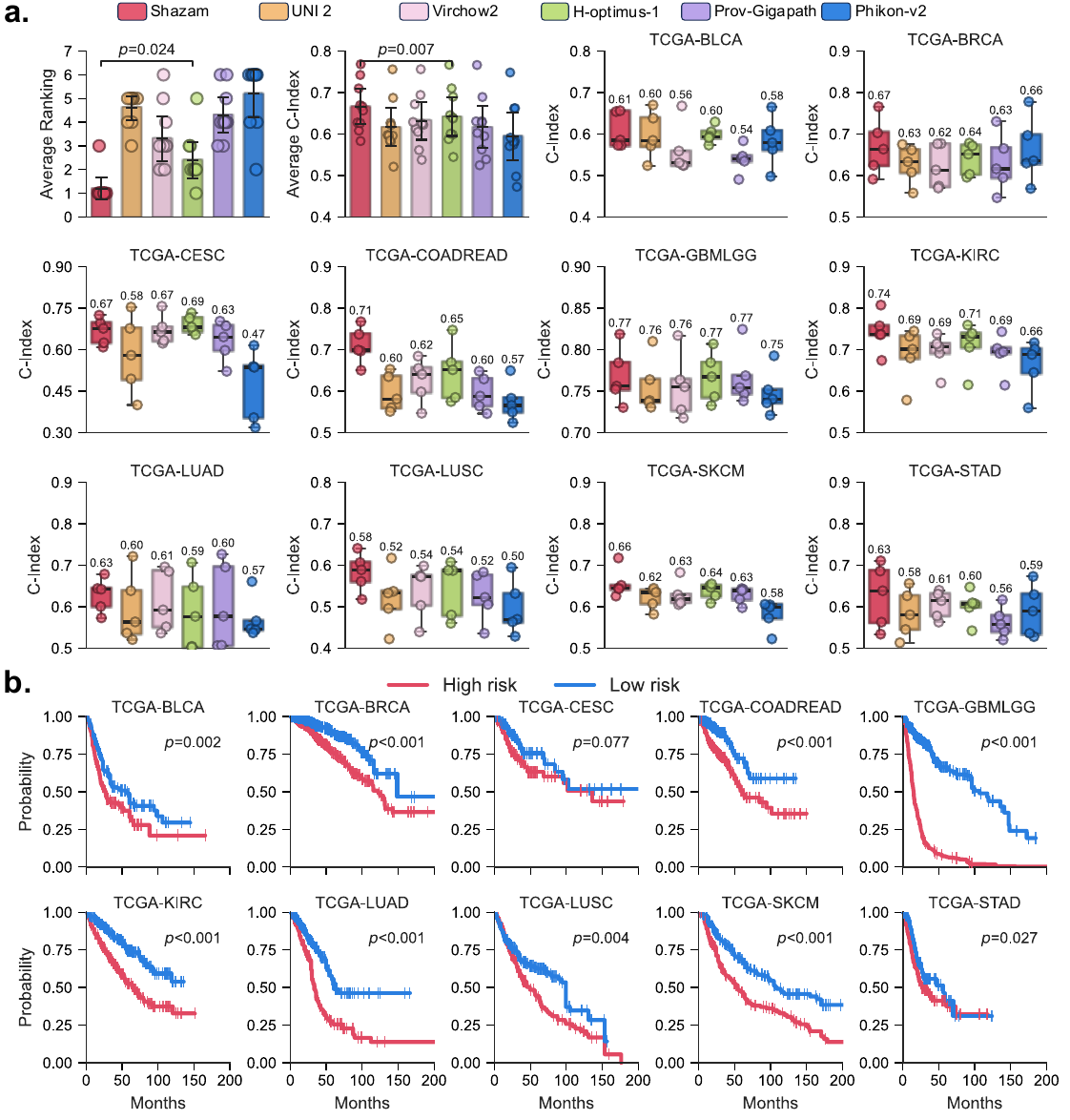}
\captionof{figure}{
\textbf{Evaluation of survival prediction across multiple cancer cohorts.}
\textbf{a}, C-index comparison of Shazam and baseline pathology foundation models across multiple cancer cohorts. Box plots show the median (center line), interquartile range (IQR, box limits represent 25th and 75th percentiles), and whiskers extending to 1.5 × IQR; outliers are shown as individual points.
\textbf{b}, Kaplan–Meier curves for a representative cohort, showing significant separation between high-risk and low-risk groups predicted by Shazam (two-sided log-rank test, $p < 0.05$), except the TCGA-CESC.}
\label{fig:fig3}

\vspace{1.0em}
cutoff. As shown in Fig.~\ref{fig:fig3}b, Shazam produced clear and statistically significant separation between survival curves in nine of the ten cancer types (two-sided log-rank test, $p < 0.05$ in all except CESC). The pronounced divergence between the high- and low-risk groups is consistent with Shazam’s elevated C-index values and demonstrates its ability to extract outcome-relevant morphologic features from routine histology. These findings collectively show that Shazam’s unified representations not only improve quantitative survival prediction but also support robust patient stratification across diverse cancer types, underscoring its potential value for downstream clinical decision support\cite{el2025whole}.

\section{Shazam improves tile-level performance across diverse histopathology tasks}
\paragraph{}Tile-level tasks provide a stringent assessment of a model’s ability to capture fine-grained morphological patterns, local microenvironmental variation, and institution- or stain-specific distribution shifts, factors that are often attenuated in whole-slide aggregation benchmarks. Rather than relying on a single foundation model with fixed inductive biases, Shazam integrates multi-scale features from multiple complementary pathology foundation models and adaptively selects informative channels at inference time. This design yields flexible tile-level representations that combine diverse morphological primitives and better accommodate heterogeneity across tissues and acquisition settings.

\paragraph{}To systematically evaluate this integrative capability, we benchmarked Shazam on eleven diverse tile-level datasets, including ESCA for esophageal carcinoma subtyping\cite{tolkach_yuri_2023_7548828}, PanCancer-TCGA for pan-cancer tissue classification\cite{komura2022universal}, PanCancer-TIL for tumor-infiltrating lymphocyte assessment\cite{saltz2018spatial}, Chaoyang for colon pathology\cite{zhu2021hard}, BACH for breast cancer categorization\cite{aresta2019bach}, CCRCC-TCGA-HEL for renal carcinoma tissue classification\cite{brummer2023computational}, CRC-MSI for microsatellite instability screening\cite{kather2019deep}, PCAM for metastatic detection\cite{bejnordi2017diagnostic}, UniToPatho for colorectal polyp grading\cite{barbano2021unitopatho}, CRC-100K\cite{kather2019predicting} and HunCRC for colorectal tissue classification\cite{pataki2022huncrc}. 

\paragraph{}As summarized in Fig. \ref{fig:fig4}a, these benchmarks collectively span a broad spectrum of tile-level task categories, including normal tissue identification, precancerous lesion grading, malignant tumor recognition, tumor microenvironment characterization, and molecular phenotype prediction. This heterogeneous task coverage enables a comprehensive and clinically representative evaluation of Shazam’s tile-level generalization performance.

\paragraph{}Across the eleven tile-level benchmarks, Shazam achieved the strongest overall performance, with a mean ranking score of 1.36 compared with 2.46 for the strongest baseline, Virchow2, underscoring the consistent advantage of its integrative multi-model formulation (Fig.~\ref{fig:fig4}a; Supplementary Tables~\ref{tab:Table7}–\ref{tab:Table17}). Shazam delivered the highest F1 score on several representative datasets. On PCAM, it achieved 0.914 and surpassed Virchow2 by approximately 1.4\%. On PanCancer-TIL, Shazam reached 0.852, outperforming Virchow2 by about 1.0\%. On PanCancer-TCGA, it attained 0.844, exceeding H-optimus-1 at 0.833 by roughly 1.1\%. These improvements illustrate how Shazam leverages complementary multi-scale features from multiple foundation models to produce more discriminative and robust tile-level descriptors across varied tissue contexts. 

\includegraphics[width=1.0\linewidth]{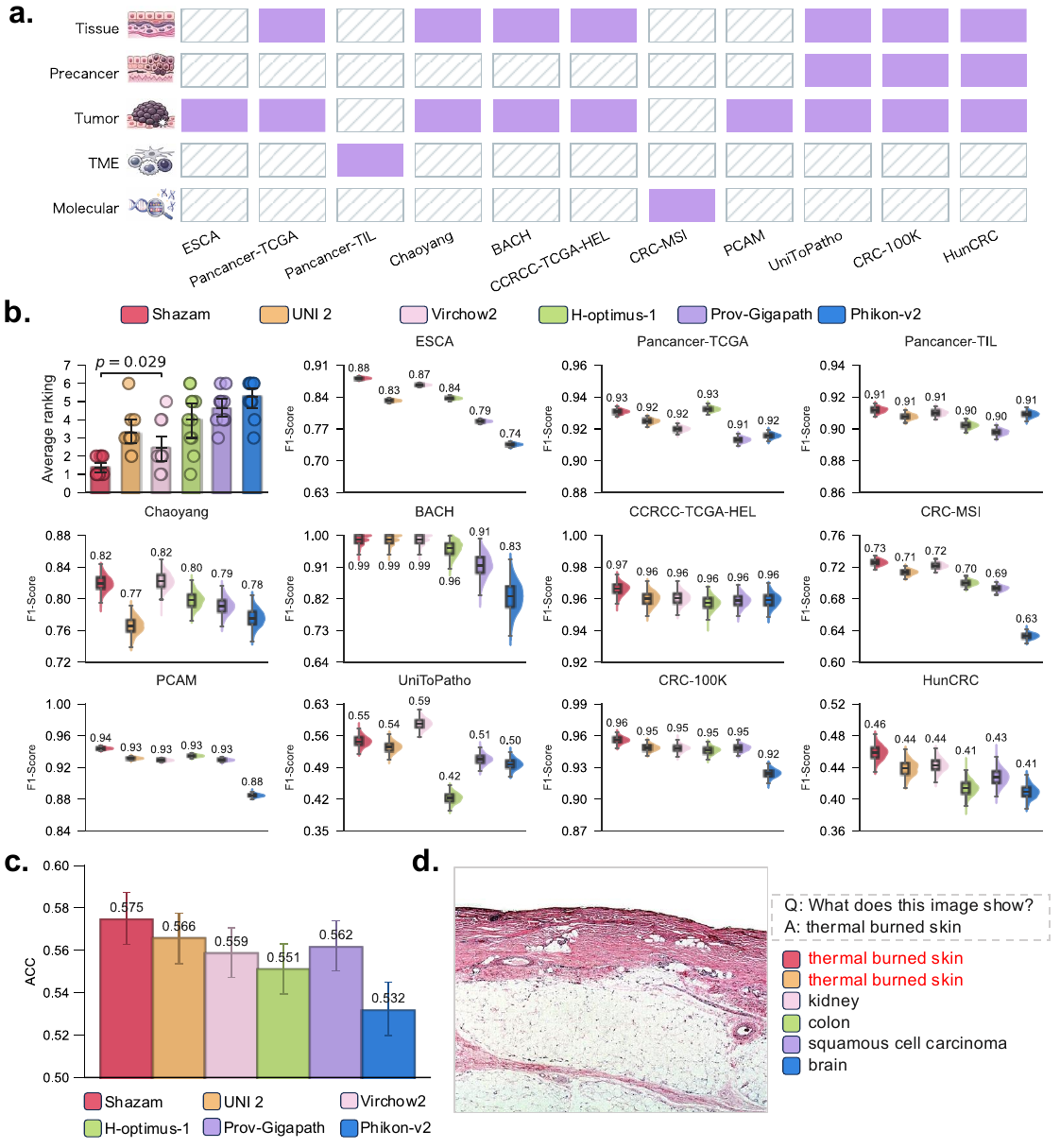}
\captionof{figure}{
\textbf{Evaluation of tile-level classification and pathology VQA.}
\textbf{a}, Coverage of tile-level task categories across benchmark datasets, with filled blocks indicating the availability of corresponding annotations.
\textbf{b}, Quantitative comparison of tile-level performance across eleven classification benchmarks measured by F1 score. The reported $p$-value is from a one-sided Wilcoxon signed-rank test comparing Shazam with the strongest individual baseline; per-dataset F1 score distributions are shown across repeated trials.
\textbf{c}, Overall VQA accuracy on the PathVQA test set, with error bars indicating 95\% confidence intervals estimated by 1000 bootstrap samples.
\textbf{d}, Example question and model-predicted answers for a representative query image.
}
\label{fig:fig4}

\vspace{1.0em}

\paragraph{} Although Shazam led performance on most benchmarks, its advantage narrowed in datasets where a single baseline model exhibited a pronounced task-specific strength. For instance, on UniToPatho, Virchow2 achieved a F1 score of 0.820, noticeably higher than the other individual models, and Shazam obtained 0.803, reflecting that the ensemble can inherit attenuated signals when several constituent models underperform. Even in such settings, Shazam remained consistently competitive across metrics, maintaining reliable performance despite variations in stain, scanner, and tissue preparation. This stability highlights Shazam’s capacity to generalize fine-grained morphological patterns at the tile level, where subtle artifacts, microenvironmental cues, and local histologic variability pose the greatest challenges for computational pathology models.

\section{Shazam improves pathology VQA}
\paragraph{}Visual Question Answering (VQA) aims to answer natural-language queries based on visual evidence and has emerged as a promising framework for assisting clinicians in retrieving task-specific diagnostic information from histopathology images. To examine Shazam’s capability in this setting, we used the PathVQA\cite{he2020pathvqa} dataset, which is among the largest resources for pathology VQA. The dataset contains 32,799 image–question–answer triplets, divided into a training set of 16,400 samples, a validation set of 9,840 samples, and a test set of 6,560 samples. Overall performance across the evaluated pathology foundation models is presented in Fig.~\ref{fig:fig4}c, with detailed results provided in Supplementary Table \ref{tab:PathVQA_top1}.

\paragraph{}Shazam achieved the highest accuracy of 0.575, exceeding the performance of the second-best model, UNI 2, by 1\%. This result is notable because the foundation models underlying Shazam were trained exclusively with image-based self-supervision and therefore received no explicit language or cross-modal guidance. Despite this limitation, the unified multi-model, multi-scale architecture of Shazam enabled strong performance on a task requiring alignment between visual patterns and linguistic queries. These findings indicate that Shazam’s integrative feature representation improves not only the discriminability of image-based features but also their transferability to cross-modal reasoning tasks. As a result, Shazam is able to extract visual cues from histopathology images that are relevant to the semantics of each question.

\paragraph{}We further provide qualitative comparisons in Fig.~\ref{fig:fig4}d, where we illustrate a representative query image, question, and predicted answers across different foundation models. Shazam and UNI 2 consistently produced more reliable and clinically meaningful responses than the other evaluated models, supporting the quantitative advantages observed in the benchmark analysis.

\includegraphics[width=1.0\linewidth]{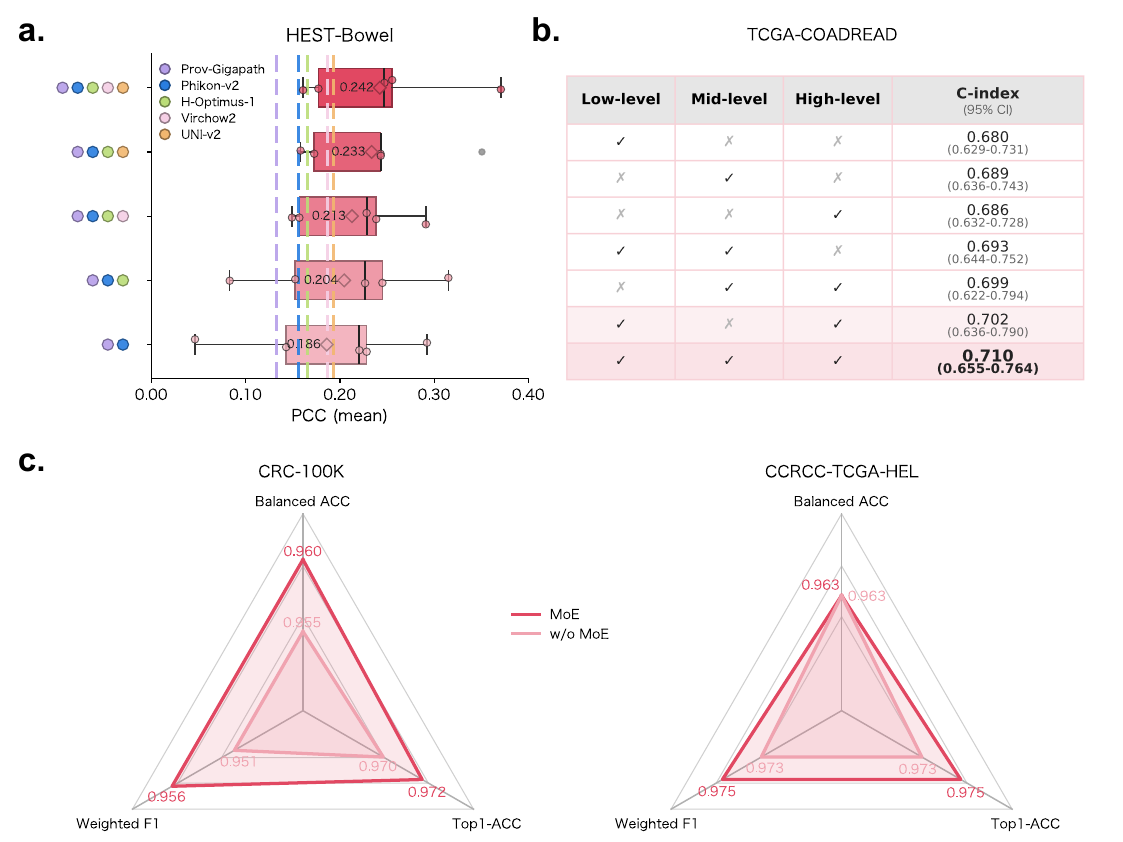}
\captionof{figure}{\textbf{Ablation analysis of Shazam.}
\textbf{a}, Effect of teacher supervision on HEST-Bowel. Box plots show the median (center line), mean (rhombus),interquartile range (IQR, box limits represent 25th and 75th percentiles), and whiskers extending to 1.5 × IQR; outliers are shown as individual points.
\textbf{b}, Complementarity of semantic feature representations for WSI-level survival prediction on TCGA-COADREAD.
\textbf{c}, Contribution of the Mixture-of-Experts (MoE) module on CRC-100K and CCRCC classification tasks.}
\label{fig:fig5}

\section{Ablation study of Shazam}

\paragraph{}
To further dissect the sources of Shazam’s performance gains, we conducted a unified ablation study examining the contributions of teacher supervision, semantic feature representations, and adaptive aggregation. These experiments are designed to quantify how performance degrades as informative signals are removed and to clarify which components contribute most to robustness across tasks.

\paragraph{}
We first analyzed the role of teacher supervision by progressively removing teacher models on the HEST-Bowel dataset (Fig.~\ref{fig:fig5}a; Supplementary Table \ref{tab:ablation_pcc_hest_bowel}). Teachers were ablated in descending order of their standalone performance. With all teachers retained, Shazam achieved a mean PCC of 0.242. Removing the strongest teacher (UNI-v2) reduced the mean PCC to 0.213, corresponding to a relative decrease of 12.0\%. In contrast, removing the second-best teacher (Virchow2) led to a smaller drop, with a mean PCC of 0.233 (a 3.7\% decrease). Together, these results support our hypothesis that prioritizing task-aligned foundation models yields more effective supervision than treating all teachers equally, and that integrating such models within Shazam leads to measurable and robust performance improvements.

\paragraph{}
We next quantified the contribution of feature representations at different levels of semantic abstraction for whole-slide survival prediction on TCGA-COADREAD (Fig.~\ref{fig:fig5}b; Supplementary Table \ref{tab:SurvivalAblation}). Using features from a single level yielded C-index values in the range of 0.680–0.689. Pairwise combinations improved performance by up to 0.022 in absolute C-index, with the low- and high-level combination achieving 0.702. Aggregating low-, mid-, and high-level features achieved the highest C-index of 0.710, representing an absolute improvement of 0.030 over the best single-level configuration and 0.008 over the best pairwise combination. These results demonstrate that representations encoding local morphology, intermediate tissue organization, and global contextual patterns capture complementary prognostic information.

\paragraph{}
Finally, we evaluated the effect of the Mixture-of-Experts (MoE) module on CRC-100K and CCRCC-TCGA-HEL classification tasks (Fig.~\ref{fig:fig5}c; Supplementary Tables~\ref{tab:TileAblationPCAM}-\ref{tab:TileAblationCCRCC}). On CRC-100K, incorporating MoE improved balanced accuracy from 0.955 to 0.960 (+0.005), weighted F1 from 0.951 to 0.956 (+0.005), and top-1 accuracy from 0.970 to 0.972 (+0.002). On CCRCC-TCGA-HEL, the corresponding improvements were smaller but consistent, with weighted F1 increasing from 0.973 to 0.975 (+0.002) and top-1 accuracy from 0.973 to 0.975 (+0.002). These results suggest that adaptive expert routing provides measurable but dataset-dependent gains over uniform feature aggregation.

\paragraph{}
Together, these ablation studies indicate that Shazam’s performance gains arise from the interaction of strong but uneven teacher supervision, complementary semantic feature representations, and adaptive aggregation. While the largest gains stem from retaining informative teachers and multi-level features, MoE further refines performance by enabling task-adaptive fusion of heterogeneous representations.

\paragraph{}

%% file: sections/3-discussion.tex
\paragraph{} In this work, we introduce Shazam, a unified pathology foundation model that consolidates the capabilities of multiple pretrained expert models without requiring access to their training data or performing any form of large-scale distillation pretraining. Previous efforts such as GPFM attempted to merge knowledge through offline distillation, yet their effectiveness is fundamentally limited by the diversity and scale of the available distillation corpus. Shazam adopts a fundamentally different integration strategy that operates in an online and task-specific manner, combining pretrained pathology foundation models through multi-level feature fusion and an adaptive expert-weighting mechanism. This design allows Shazam to continuously benefit from advances in pathology foundation models and to flexibly incorporate the most recent and most powerful models for any specific downstream task. With this ability, Shazam demonstrates consistently strong performance across four major categories of computational pathology, including spatial transcriptomics prediction, survival analysis, tile-level classification, and visual question answering in pathology. Across a total of 30 benchmark tasks, Shazam surpasses leading individual models such as UNI 2 and Virchow2 in the majority of evaluations.

\paragraph{} The rapid development of pathology foundation models in recent years has markedly advanced the field, yet their pretraining pipelines continue to rely heavily on large, privately curated datasets whose composition varies substantially across institutions. These differences lead individual models to specialize in particular tissue types, staining styles or diagnostic tasks, which makes it difficult to ensure broad generalization across the diverse landscape of computational pathology. Moreover, because the underlying training data of these models are inaccessible, progress in foundation model development cannot accumulate at the data level, and each new model is effectively trained in isolation from its predecessors. Shazam addresses this structural limitation by allowing downstream tasks to selectively integrate the strengths of multiple pretrained models, without requiring access to any of their training corpora. Through this design, Shazam demonstrates robust generalization across heterogeneous tasks, suggesting that online integration of multiple foundation models represents a promising direction for the next generation of computational pathology systems. This is particularly valuable in settings where different models capture distinct aspects of morphological or molecular variation, and where a single architecture is unlikely to dominate across all task domains.

\paragraph{} Despite its strong empirical performance, Shazam has several limitations that warrant further investigation. Because Shazam requires all constituent foundation models to perform feature extraction during both training and inference, its computational cost grows approximately linearly with the number of integrated models. This requirement may limit its practicality in settings where low latency or constrained hardware resources are essential. Recent progress in developing lightweight pathology foundation models with substantially fewer parameters yet competitive performance provides a promising avenue for reducing this overhead\cite{filiot2025distilling}, and future versions of Shazam could incorporate such models to achieve a more favorable efficiency–performance balance. Another limitation lies in the breadth of external validation. Although Shazam demonstrates consistent gains across a wide range of internal benchmarks, the evaluation on independent datasets remains limited. Expanding external validation to additional cohorts and institutions will be important for fully establishing its robustness under real-world distributional variability. Continued exploration along these directions may further enhance the scalability, generalizability, and clinical readiness of Shazam.

\paragraph{}

%% file: supplement/1-framework.tex
\setcounter{section}{0}
\heading{Framework}
\section{Shazam}
Shazam is a multi-scale fusion framework designed to synthesize visual representations from diverse pathology foundation models. 
It consists of four key components: (1) multi-teacher and multi-scale representation extraction, (2) Feature alignment and scale-wise Fusion, (3) Patch-to-slide aggregation for WSI tasks, and (4) Distillation across scales and teachers. 
Rather than relying on a single pathology foundation model, Shazam leverages complementary information distributed across different architectures and across multiple semantic scales.
\paragraph{Multi-teacher and multi-scale representation extraction.}
For each foundation model, we extract visual features at three characteristic depths—early, middle, and final layers—corresponding to low-, mid-, and high-level semantic representations. 
To ensure reproducibility across diverse architectures, we determine the extraction points mathematically. Let $L$ denote the total number of transformer blocks in a given teacher model. We place non-invasive forward hooks at block indices defined by $l_{\text{low}} = \lfloor 0.33 \times L \rfloor$, $l_{\text{mid}} = \lfloor 0.66 \times L \rfloor$, and $l_{\text{high}} = L$.
During the training of Shazam, all teacher models remain frozen to serve as stable feature extractors.
This strategy offers a consistent and lightweight mechanism for capturing multi-scale information across the five foundation models used in this study, namely UNI 2\cite{chen2024towards}, Virchow2\cite{zimmermann2024virchow2}, H-optimus-1\cite{hoptimus1}, Prov-Gigapath\cite{xu2024whole}, and Phikon-v2\cite{filiot2024phikonv2largepublicfeature}.

\paragraph{Feature alignment and scale-wise Fusion.} 
In Shazam, the multi-scale features from multiple teacher models are first aligned into a shared embedding space to enable effective fusion. Given the multi-scale features from different foundation models, we concatenate the features across all experts and apply a gating mechanism to weight the contribution of each teacher's representation. Specifically, let \(\mathbf{f}_1, \mathbf{f}_2, \dots, \mathbf{f}_N\) denote the feature representations from \(N\) teacher models, where each \(\mathbf{f}_i \in \mathbb{R}^{d_{\text{model}}}\) is the feature from the \(i\)-th teacher. The features from each teacher model are concatenated into a single vector:
\[
\mathbf{f}_{\text{concat}} = \text{concat}(\mathbf{f}_1, \mathbf{f}_2, \dots, \mathbf{f}_N) \in \mathbb{R}^{N \cdot d_{\text{model}}}
\]
A gating network is applied to the concatenated feature vector to compute the contribution weights for each expert. The gating mechanism consists of a simple multilayer perceptron (MLP) that takes the concatenated features and produces the gating scores, \( \mathbf{g} \). Specifically, the gating scores are computed as:
\[
\mathbf{g} = \text{softmax}\left(\text{MLP}(\mathbf{f}_{\text{concat}})\right)
\]
where \( \mathbf{f}_{\text{concat}} \) is the concatenated feature vector from all teachers, and \( \text{MLP} \) represents a multi-layer perceptron. The output, \( \mathbf{g} \), is a set of weights that determine the contribution of each expert's feature to the final fused representation.

The weighted feature from each teacher is first computed as \( g_i \mathbf{f}_i \), where \( g_i \) denotes the gating score for the \(i\)-th teacher and \( \mathbf{f}_i \in \mathbb{R}^d \) is the corresponding feature representation. Rather than concatenating these weighted features into a single vector, the Mixture-of-Experts (MoE) module organizes them into a feature matrix, with each row encoding the gated contribution of one teacher. Formally, the fused representation is defined as
\[
\mathbf{F}_{\text{fused}}
=
\begin{bmatrix}
g_1 \mathbf{f}_1^\top \\
g_2 \mathbf{f}_2^\top \\
\vdots \\
g_N \mathbf{f}_N^\top
\end{bmatrix}
\in \mathbb{R}^{N \times d},
\]
where \(N\) is the number of teacher models and each row corresponds to a teacher-specific, gated feature vector. This fused feature matrix preserves the individual contributions of different teachers and is subsequently passed to the self-attention layers, which model interactions across teachers and adaptively aggregate their information for downstream prediction.

To further enhance the quality of the fused representation, we apply a small stack of self-attention blocks to the stacked weighted features. These blocks facilitate information exchange across features at different semantic scales, improving the model's ability to capture both fine-grained details and high-level context. The self-attention mechanism uses the stacked weighted features as input and learns to focus on important interactions between the features:
\[
\mathbf{f}_{\text{att}} = \text{SelfAttention} (\mathbf{f}_{\text{fused}})
\]
Finally, the output from the self-attention block is passed through a layer normalization \cite{ba2016layer} step to stabilize the final feature representation:
\[
\mathbf{f}_{\text{final}} = \text{LayerNorm} (\mathbf{f}_{\text{att}})
\]
This multi-scale, expert-weighted fusion process enables Shazam to capture complementary information from multiple teacher models, facilitating effective feature alignment and fusion across scales.

\paragraph{Patch-to-slide aggregation for WSI tasks.}
For whole-slide image (WSI) analysis, each teacher produces patch-level features at all three scales. 
These patch embeddings are aggregated into WSI-level representations using the Attention-Based Multiple Instance Learning (ABMIL) mechanism proposed by Ilse et al.\cite{ilse2018attention}.
Shazam employs scale-specific, independent ABMIL heads to aggregate features at each of the low, mid, and high scales, preserving hierarchical distinctiveness before the final fusion and prediction. Tile-level tasks directly use per-image features without patch aggregation.

\paragraph{Distillation across scales and teachers.}
To ensure that the fused representation retains the complementary strengths of individual foundation models, Shazam employs a multi-level distillation objective across three semantic scales.
For each scale $s \in \{\mathrm{low}, \mathrm{mid}, \mathrm{high}\}$, we $\ell_2$-normalize both the student feature $z_s$ and the $i$-th teacher feature $t_s^{(i)}$ prior to distillation to stabilize cosine-based alignment.
The student is then supervised to match each teacher using a combination of cosine distance and an element-wise Huber loss:
\begin{equation}
\mathcal{L}_{\mathrm{distill}}^{(s,i)}
=
\left(
1 - \frac{\langle z_s,\, t_s^{(i)} \rangle}{\|z_s\|\,\|t_s^{(i)}\|}
\right)
+
\mathrm{Huber}\!\left(z_s,\, t_s^{(i)}\right),
\end{equation}
where $\langle \cdot,\cdot\rangle$ denotes the inner product and $\|\cdot\|$ denotes the Euclidean norm.

\noindent\textbf{Element-wise Huber loss.}
Let $e_{s,k}^{(i)} = z_{s,k} - t_{s,k}^{(i)}$ be the discrepancy at feature dimension $k \in \{1,\ldots,d\}$.
We define the element-wise Huber loss as
\begin{equation}
\mathrm{Huber}\!\left(z_s, t_s^{(i)}\right)
=
\frac{1}{d}\sum_{k=1}^{d}
\begin{cases}
\frac{1}{2}\left(e_{s,k}^{(i)}\right)^2, & \left|e_{s,k}^{(i)}\right|\le \delta, \\[4pt]
\delta\left(\left|e_{s,k}^{(i)}\right|-\frac{1}{2}\delta\right), & \left|e_{s,k}^{(i)}\right|>\delta,
\end{cases}
\end{equation}
where $\delta>0$ is the transition threshold and $d$ is the feature dimension.
This formulation is quadratic for small discrepancies and linear for larger deviations, reducing sensitivity to outlier channels in teacher features.

Finally, we average the distillation objective over all scales and all $N=5$ teacher models:
\begin{equation}
\mathcal{L}_{\mathrm{distill}}
=
\frac{1}{3N}
\sum_{s \in \{\mathrm{low}, \mathrm{mid}, \mathrm{high}\}}
\sum_{i=1}^{N}
\mathcal{L}_{\mathrm{distill}}^{(s,i)}.
\end{equation}

\paragraph{Training objective.}
Shazam is trained jointly with task-specific supervision signals such as cross-entropy for classification or negative log-likelihood for survival analysis. The final objective combines the task-level loss ($\mathcal{L}_{\text{task}}$) and the multi-scale distillation loss ($\mathcal{L}_{\text{distill}}$):
\[
\mathcal{L}
= \mathcal{L}_{\text{task}}
+ \lambda_{\text{distill}}\, \mathcal{L}_{\text{distill}},
\]
where $\lambda_{\text{distill}} = 0.01$ controls the strength of the distillation constraint. This formulation allows Shazam to adapt to individual downstream tasks while maintaining consistency with the representational structures encoded by the frozen teacher models. 

For optimization, we use the AdamW optimizer with a weight decay of $1 \times 10^{-4}$ and employ a cosine learning rate schedule. The model is trained for 50 epochs with a batch size of 128.

Overall, Shazam provides a unified representation-learning module capable of harmonizing information from multiple pathology foundation models and across several semantic scales, enabling improved robustness and transferability across diverse computational pathology applications.

\section{Comparisons \& Baselines}

\paragraph{}To ensure a fair and controlled evaluation, all baseline pathology foundation models were assessed using the same downstream pipelines as Shazam. Each baseline was used strictly as a frozen feature extractor according to its official implementation, and all downstream components were kept identical across methods. This includes the task-specific prediction heads as well as the ABMIL module for whole-slide analysis. Patch embeddings were consistently taken from the final transformer block of each baseline model. By standardizing the feature extraction procedure, the aggregation architecture, and the optimization settings, the comparisons isolate the representational differences among models and ensure that performance variation reflects the intrinsic quality of the learned features.

%% file: supplement/2-datasets.tex
\heading{Tasks and datasets}

\section{Spatial Transcriptomic Experiments}
\paragraph{}Spatial transcriptomics measures gene expression while preserving the anatomical location of each measurement within intact tissue sections \cite{staahl2016visualization}. By linking RNA abundance directly to histological context, these technologies enable spatially resolved characterization of cell states, signaling programs, and tumor–microenvironment interactions that cannot be inferred from morphology or bulk profiling alone \cite{staahl2016visualization,pineiro2022research}. This integration of molecular and histologic information provides a powerful framework for understanding tissue organization and disease progression.

\paragraph{}We conducted experiments on the publicly available HEST-1k dataset \cite{jaume2024hest}, which collected paired spatial transcriptomic profiles and matched H\&E-stained whole-slide images across multiple organs. Cohorts were derived from the dataset metadata (CSV) by restricting to malignant cases (disease\_state = cancer) and selecting organ labels in {breast, prostate, skin, bowel, liver, kidney, brain, bladder}, yielding eight organ-specific subsets with sufficient paired samples. To avoid patient-level leakage, we used a patient-stratified split: samples lacking patient identifiers were assigned to the training set only, whereas samples with identifiers were partitioned at the patient level such that all associated spots and image patches from a given patient appeared in a single fold.

\paragraph{}To define clinically meaningful prediction targets, we curated cohort-specific gene lists guided by FDA-recognized biomarkers in OncoKB \cite{oncokb_fda_recognition}. Cohort-level preprocessing was then applied to reduce sparsity and ensure adequate expression signal. Slides were excluded if at least half of the selected genes had zero expression, and genes were excluded if they were zero in at least half of the slides within the same cohort. The resulting filtered datasets were used for all downstream analyses. Cohort-specific gene lists are summarized in Supplementary Table~\ref{tab:cancer_gene_lists_csv_1} and Supplementary Table~\ref{tab:cancer_gene_lists_csv_2}.

\paragraph{}Model inputs were constructed by extracting H\&E patches of $112\times112\ \mu\mathrm{m}$, corresponding to 224$\times$224 pixels at 20$\times$ magnification, paired with their matched gene expression vectors. Models were trained using patient-level $k$-fold cross-validation with $k = \min(5, N_\text{patients})$. Prior to training, gene expression was cohort-wise log-normalized to stabilize variance. The student regressor consisted of a deep MLP; expert features were concatenated and fused using stacked cross-attention layers, followed by an MLP with GELU activations and dropout, and a final linear output layer predicting all target genes.Optimization used AdamW with a learning rate of $1\times10^{-3}$ for 30 epochs and a validation-driven scheduler. The loss combined ridge regression (mean squared error with $L_2$ regularization) with a multi-level distillation term, forming $\mathcal{L}=\mathcal{L}_{\mathrm{ridge}}+\lambda_{\mathrm{distill}}\mathcal{L}_{\mathrm{distill}}$, where $\lambda_{\mathrm{distill}}=0.01$.

\section{Survival Analysis}
In oncology, prognostic modeling aims to link patient time-to-event outcomes (e.g., progression and overall survival) to measurable evidence. By providing a wealth of tissue-scale data, whole-slide images allow survival modeling to take advantage of histomorphological cues and connect them to clinical prognosis.

To preprocess whole-slide images (WSIs), we adopt the CLAM pipeline~\cite{lu2021data} to perform tissue foreground segmentation and extract valid regions. From these foreground masks, we crop non-overlapping patches at 20$\times$ magnification and resize each patch to 224$\times$224 pixels for feature extraction. Slides containing too few valid patches after segmentation are excluded. The resulting patch set for each WSI is then passed to pathology foundation models to obtain patch-level embeddings for subsequent MIL-based slide analysis. 

In this study, we adopt the Attention-Based Multiple Instance Learning (ABMIL) framework \cite{ilse2018attention} for survival analysis, employing the Negative Log-Likelihood (NLL) loss. we introduce a distillation loss term for model compression. Specifically, we employ Shazam student model, distilled from a larger teacher model, using a weighted distillation loss with $\lambda = 0.01$ to align the student’s predictions with the teacher’s. Optimization is performed using AdamW with a learning rate of \(2 \times 10^{-4}\) and weight decay of \(1 \times 10^{-3}\).The dataset partitioning strategy follows that described in the GPFM study \cite{ma2024towards}. To assess the impact of different pathology foundation models on survival prediction, we perform 5-fold cross-validation and use the C-index as the evaluation metric.  

Experiments are conducted on 10 TCGA datasets: TCGA-BRCA, TCGA-BLCA, TCGA-CESC, TCGA-COADREAD, TCGA-GBMLGG, TCGA-KIRC, TCGA-LUAD, TCGA-LUSC, TCGA-SKCM and TCGA-STAD. The results are reported in Supplementary Table~\ref{tab:Table5}-~\ref{tab:Table7}.
\subsubsection*{Breast Cancer Survival Prediction.}
In the breast cancer survival prediction task, the TCGA-BRCA dataset~\cite{tcga-dataset} is adopted, containing 1,023 cases with a total of 1,089 WSIs. Following the GPFM partitioning strategy, we stratify survival times into four equally populated bins for both training and testing to ensure comparable censoring patterns across splits. For each fold of the 5-fold cross-validation, we perform an 8:2 label-stratified split, resulting in 821 training cases and 202 testing cases. The results across the 5 folds are presented in Supplementary Table \ref{tab:Table5}.

\subsubsection*{Bladder Cancer Survival Prediction.}
For bladder cancer survival prediction, we use the TCGA-BLCA dataset\cite{tcga-dataset}, containing 376 cases (446 WSIs). Using the same stratified binning and partitioning strategy as BRCA, each fold allocates 305 cases to training and 71 cases to testing. The results across the 5 folds are presented in Supplementary Table \ref{tab:Table5}.

\subsubsection*{Cervical Squamous Cell Carcinoma and Endocervical Adenocarcinoma Survival Prediction.}
The TCGA-CESC\cite{tcga-dataset} dataset includes 250 cases (260 WSIs). Using the same strategy, we assign 203 cases to training and 47 cases to testing per fold. The results across the 5 folds are presented in Supplementary Table \ref{tab:Table5}.

\subsubsection*{Colon \& Rectum Adenocarcinoma Survival Prediction.}
We combine TCGA-COAD\cite{tcga-dataset} (426 cases, 434 WSIs) and TCGA-READ (153 cases, 154 WSIs) into a single dataset of 579 cases (588 WSIs). Each fold assigns 464 cases to training and 115 to testing. The results across the 5 folds are presented in Supplementary Table \ref{tab:Table6}.

\subsubsection*{Glioma Survival Prediction.}
We merge TCGA-GBM (370 cases, 823 WSIs) and TCGA-LGG\cite{tcga-dataset} (460 cases, 778 WSIs) into a combined glioma dataset of 830 cases (1,601 WSIs). Per fold, 667 cases are used for training and 163 for testing. The results across the 5 folds are presented in Supplementary Table \ref{tab:Table6}.

\subsubsection*{Kidney Renal Clear Cell Carcinoma Survival Prediction.}
The TCGA-KIRC dataset\cite{tcga-dataset} comprises 498 cases (504 WSIs). Following the same partitioning strategy, each fold assigns 401 cases to training and 97 cases to testing. The results across the 5 folds are presented in Supplementary Table \ref{tab:Table6}.

\subsubsection*{Lung Adenocarcinoma Survival Prediction.}
For TCGA-LUAD\cite{tcga-dataset}, we use 455 cases (518 WSIs), splitting each fold into 366 training cases and 89 testing cases. The results across the 5 folds are presented in Supplementary Table \ref{tab:Table7}.
\subsubsection*{Lung Squamous Cell Carcinoma Survival Prediction.}
The TCGA-LUSC dataset\cite{tcga-dataset} comprises 452 cases (484 WSIs). Using the same stratified splitting approach, 365 cases are assigned to training and 87 to testing in each fold. The results across the 5 folds are presented in Supplementary Table \ref{tab:Table7}.

\subsubsection*{Skin Cutaneous Melanoma Survival Prediction.}
The TCGA-SKCM dataset\cite{tcga-dataset} contains 415 cases (456 WSIs). Each fold uses 337 cases for training and 78 for testing. The results across the 5 folds are presented in Supplementary Table \ref{tab:Table7}.

\subsubsection*{Stomach Adenocarcinoma Survival Prediction.}
The TCGA-STAD dataset\cite{tcga-dataset} contains 363 cases (389 WSIs). In each fold, 293 cases are allocated to training and 70 to testing using the same stratified splitting procedure. The results across the 5 folds are presented in Supplementary Table \ref{tab:Table6}.

\section{Tile-level Classification}

\paragraph{}For each foundation model, we load its official pretrained weights and extract tile-level representations following the recommended preprocessing settings. This ensures that all models are evaluated under a consistent and standardized protocol.

\paragraph{}Based on these extracted features, we train a multilayer perceptron (MLP) classifier composed of two fully connected layers with LayerNorm and GELU activation, followed by a final linear prediction layer. Optimization is performed using Adam with a learning rate of \(1\times 10^{-3}\), a batch size of 64, and a maximum of 100 epochs with early stopping (patience = 30). This procedure is used to assess the representational quality and transferability of each foundation model when used in isolation.

\paragraph{}For our method, Shazam, tile-level features are first extracted from all five foundation models at three semantic scales. These multi-scale features are subsequently fused through the Shazam framework: each scale is processed by a Mixture-of-Experts module to adaptively combine information across models, followed by layers of self-attention to refine cross-model interactions. The resulting fused representation is then passed to the same MLP classifier used for the baselines. During training, we additionally apply a multi-level distillation objective that aligns the fused Shazam feature at each scale with its corresponding teacher-specific representation, encouraging the fused embedding to inherit complementary signals from all foundation models.

\paragraph{}The overall training objective is defined as:
\[
\mathcal{L} = \mathcal{L}_{\mathrm{CE}} \;+\; \lambda_{\mathrm{distill}} \,\mathcal{L}_{\mathrm{distill}},
\]
where \(\lambda_{\mathrm{distill}} = 0.01\). The distillation term is composed of a cosine-similarity loss and a Huber loss applied at each semantic scale. To properly handle class imbalance, we report Balanced Accuracy, Weighted F1, and Top1-ACC as the evaluation metrics.

\subsubsection*{HunCRC for CRC Tissue Classification (9 classes).}
This dataset comprises 101,398 H\&E-stained image patches, also referred to as regions of interest (ROIs), each measuring 512 × 512 pixels at a resolution of 0.48 microns per pixel (mpp). These ROIs were extracted from 200 formalin-fixed paraffin-embedded (FFPE) whole-slide images (WSIs) of colorectal biopsies\cite{pataki2022huncrc}. Designed to support the training of machine learning models for colorectal tissue classification, the dataset includes annotations for nine classes: Adenocarcinoma (4,315 ROIs), High-Grade Dysplasia (2,281 ROIs), Low-Grade Dysplasia (55,787 ROIs), Inflammation (763 ROIs), Tumor Necrosis (365 ROIs), Suspicious for Invasion (570 ROIs), Resection Edge (534 ROIs), Technical Artifacts (3,470 ROIs), and Normal Tissue (31,323 ROIs). For training and evaluation, the dataset follows the official split, with 76,753 ROIs used for training, 11,327 ROIs for validation, and 11,328 ROIs for testing. The experimental results for this dataset are reported in Supplementary Table~\ref{tab:Table8}.

\subsubsection*{UniToPatho for CRC Polyp Classification (6 classes).}
This dataset comprises 9,536 H\&E-stained image patches (ROIs) extracted from 292 whole-slide images (WSIs)\cite{barbano2021unitopatho}. Its primary purpose is to facilitate the training of deep neural networks for colorectal polyp classification and adenoma grading. The dataset includes annotations for six classes: Normal Tissue (950 ROIs), Hyperplastic Polyp (545 ROIs), Tubular Adenoma with High-Grade Dysplasia (454 ROIs), Tubular Adenoma with Low-Grade Dysplasia (3,618 ROIs), Tubulo-Villous Adenoma with High-Grade Dysplasia (916 ROIs), and Tubulo-Villous Adenoma with Low-Grade Dysplasia (2,186 ROIs). For training and evaluation, the dataset follows the official split, with 6,270 ROIs used for training, 1,199 ROIs for validation, and 1,200 ROIs for testing. The experimental results for this dataset are reported in Supplementary Table~\ref{tab:Table9}.

\subsubsection*{PCAM for Metastatic Tissue Classification (2 classes).}
Derived from the CAMELYON16 \cite{bejnordi2017diagnostic}\cite{veeling2018rotation}, the PCAM dataset comprises 327,680 color histopathology patches, each with a native resolution of $96 \times 96$ pixels. The dataset poses a binary classification problem, where each image is labeled according to the presence of metastatic tissue. In our experiments, we adhere to the standard data partition, utilizing 262,144 images for training and 32,768 images each for validation and testing. Prior to model input, all regions of interest (ROIs) are upscaled to $224 \times 224$ pixels. The experimental results for this dataset are reported in Supplementary Table~\ref{tab:Table10}.

\subsubsection*{ESCA for Esophageal Carcinoma Subtyping (11 classes).}
ESCA contains 367,229 patches, each with a resolution of $256 \times 256$ pixels, sourced from various institutions: University Hospital Cologne (UKK), Landesklinikum Wiener Neustadt (WNS), TCGA, and University Hospital Berlin Charité (CHA)\cite{tolkach2023artificial}. These patches were annotated into 11 classes, including adventitia, lamina propria mucosae, muscularis mucosae, muscularis propria, regression tissue, mucosa gastric, mucosa esophagus, submucosa, submucosal glands, tumor, and ulceration. For training and evaluation, we use the CHA dataset (178,187 ROIs) as the training set and combine UKK, WNS, and TCGA (189,142 ROIs) for testing. All images were resized to $224 \times 224$ pixels. The experimental results for this dataset are reported in Supplementary Table~\ref{tab:Table11}.

\subsubsection*{PanCancer-TIL for TIL classification (2 classes).}
PanCancer-TIL consists of 304,097 images with a resolution of $100 \times 100$ pixels, labeled as TIL-positive (54,910 ROIs) and TIL-negative (249,187 ROIs)\cite{saltz2018spatial}. The official train-validation-test split of 209,221, 38,601, and 56,275 ROIs is adopted, and all images are resized to $256 \times 256$ pixels. The experimental results for this dataset are reported in Supplementary Table~\ref{tab:Table12}.

\subsubsection*{CRC-100K for Colorectal Cancer (CRC) Tissue Classification (9 classes).}
The CRC-100K dataset is composed of the NCT-CRC-HE-100K and CRC-VAL-HE-7K cohorts~\cite{kather2019predicting}. 
NCT-CRC-HE-100K contains 100{,}000 non-overlapping $224 \times 224$ H\&E patches sourced from 86 colorectal cancer whole-slide images obtained from the NCT biobank and the UMM pathology archive. 
The CRC-VAL-HE-7K subset includes 7{,}180 $224 \times 224$ patches collected from 50 patients diagnosed with colorectal adenocarcinoma. 
The dataset covers nine histological tissue categories: adipose (ADI, 11{,}745 ROIs), background (BACK, 11{,}413), debris (DEB, 11{,}851), lymphocytes (LYM, 12{,}191), mucus (MUC, 9{,}931), smooth muscle (MUS, 14{,}128), normal colon mucosa (NORM, 9{,}504), cancer-associated stroma (STR, 10{,}867), and colorectal adenocarcinoma epithelium (TUM, 15{,}550). 
We follow the official train and test split of 100{,}000 and 7{,}180 patches, respectively. The experimental results for this dataset are reported in Supplementary Table~\ref{tab:Table13}.

\subsubsection*{CCRCC-TCGA-HEL for CCRCC Tissue Classification (4 classes).}
We utilize a dataset of 52,713 regions of interest (ROIs) sourced from the TCGA-KIRC and Helsinki cohorts~\cite{brummer2023computational}, with each patch standardized to $300 \times 300$ pixels. While the original dataset spans six categories, we restrict our analysis to four clinically relevant classes: renal cancer (13,057), normal renal tissue (8,652), stroma (5,460), and red blood cells (996). After excluding the 'empty' (16,026) and 'other' (8,522) categories, the remaining 28,165 samples are shuffled and partitioned into a training set of 22,530 and a test set of 5,635. The experimental results for this dataset are reported in Supplementary Table~\ref{tab:Table14}.

\subsubsection*{BACH for Breast Cancer Tissue Classification (4 classes).}
The BACH dataset \cite{aresta2019bach} comprises 400 high-resolution histology images ($2048 \times 1536$), evenly distributed across four classes: Normal, Benign, In situ carcinoma, and Invasive carcinoma (100 samples per class). To facilitate model training, we resize all images to $224 \times 224$ pixels and employ a stratified random split, allocating 320 images for training and 80 for evaluation. The experimental results for this dataset are reported in Supplementary Table~\ref{tab:Table15}.

\subsubsection*{CRC-MSI for MSI Screening (2 classes).}
The CRC-MSI dataset\cite{kather2019deep} consists of 51,918 histological patches ($512 \times 512$ pixels) derived from the TCGA colorectal cancer cohort.  The data is categorized by Microsatellite Instability (MSI) status, distinguishing between patients with high instability (MSI-H) and those without (Non-MSI-H, comprising MSI-L and MSS). We adhere to the official data partition, utilizing 19,557 images for training and 32,361 for testing. The experimental results for this dataset are reported in Supplementary Table~\ref{tab:Table16}.

\subsubsection*{PanCancer-TCGA for Tissue Classification (32 classes).}
The PanCancer-TCGA dataset \cite{komura2022universal} is a large-scale collection of 271,170 histological patches ($256 \times 256$ pixels) derived from 8,736 Whole Slide Images (WSIs) in the TCGA archive.  Spanning 32 distinct cancer subtypes, the dataset includes major classes such as Head and Neck Squamous Cell Carcinoma (11,790 ROIs), Bladder Urothelial Carcinoma (9,990 ROIs), Lung Squamous Cell Carcinoma (16,560 ROIs), and Glioblastoma Multiforme (23,740 ROIs). For training and evaluation, we use a 21,736:54,342 train/test split. The experimental results for this dataset are reported in Supplementary Table~\ref{tab:Table17}.

\subsubsection*{Chaoyang for Colon Tissue Classification (4 classes).}
Sourced from Chaoyang Hospital \cite{zhu2021hard}, the Chaoyang dataset consists of 6,160 colon tissue tiles categorized into four distinct classes: Normal (1,816), Serrated (1,163), Adenocarcinoma (2,244), and Adenoma (937).  We adhere to the standard evaluation protocol, resizing all images to $224 \times 224$ pixels and employing the official training and testing partition of 4,021 and 2,139 samples, respectively. The experimental results for this dataset are reported in Supplementary Table~\ref{tab:Table18}.

\section{Pathology VQA}

We evaluate our pathology foundation models on the PathVQA benchmark~\cite{he2020pathvqa}, a standard dataset for assessing visual question answering (VQA) capabilities in pathology. PathVQA requires models to jointly interpret visual content and natural-language questions, providing a comprehensive assessment of multimodal medical reasoning. We adopt the Multi-modal Unified Medical Captioning (MUMC) framework~\cite{mumc}, which demonstrates strong performance on PathVQA by tightly integrating textual and visual representations.

The VQA model architecture consists of four main components: the image encoder, text encoder, multimodal encoder, and answering decoder. The only architectural modification we introduce lies in the image encoding stage. Instead of the single ViT encoder used in the original MUMC design, we integrate Shazam as a multi-level visual expert. Shazam provides hierarchical low-, mid-, and high-level pathology representations, which are further combined with features extracted from four additional pathology foundation models. These multi-model and multi-level visual features are fused through a Mixture-of-Experts mechanism followed by a sequence of cross-attention layers. The resulting fused embedding is projected into a 768-dimensional space to match the input requirements of the subsequent multimodal Transformer.

All remaining components of the architecture strictly follow the original MUMC configuration. The text encoder processes each question using the first six layers of a pre-trained BERT model, functioning as a pure language encoder. The multimodal encoder corresponds to the final six layers of the same BERT model, each incorporating cross-attention to fuse the projected visual embeddings with the intermediate textual representations. This two-stage design allows the model to first capture the linguistic structure of the question before grounding it in visual evidence drawn from Shazam and the other pathology foundation models. The answering decoder, implemented as a 6-layer Transformer language model, autoregressively generates answer tokens conditioned on the multimodal representations.

During training, all Shazam parameters remain frozen to preserve its domain-specific visual representation. The model is fine-tuned for 40 epochs using AdamW, with the learning rate decayed from $2\times10^{-5}$ to $1\times10^{-8}$ to ensure stable optimization. The training objective includes a standard cross-entropy loss for answer generation, combined with a hierarchical distillation loss that supervises the fused MoE representations using multi-level teacher features from the five foundation models. This distillation strategy encourages the student model to align its fused representation with the teacher experts at the low-, mid-, and high-level semantic hierarchies, further improving cross-modal alignment and VQA performance.

The experimental results are summarized in Supplementary Table~\ref{tab:PathVQA_top1}. Integrating Shazam as a multi-level visual expert leads to a substantial improvement in accuracy, demonstrating the effectiveness of hierarchical feature integration for supporting fine-grained multimodal reasoning in pathology

\section{Ablation Study}

\subsection{Spatial Transcriptomic Ablation Study}
We conducted an ablation study under 5-fold cross-validation by sequentially removing teacher models from the ensemble according to their standalone performance ranking. We first established an ordered list of teacher feature extractors based on single-teacher validation results, then initialized the ablation sequence from the complete teacher set used in Shazam and generated a series of reduced-teacher configurations by removing the top-ranked teacher, followed by the next-ranked teacher, iteratively until only a minimal subset remained. Throughout all ablation runs, we held the data splits, model architecture, training pipeline, and evaluation protocol fixed to ensure that observed differences across configurations can be attributed solely to the removal of specific teachers. The results are reported in Supplementary Table~\ref{tab:ablation_pcc_hest_bowel}.

\subsection{Survival Analysis Ablation Study}

For WSI-level survival prediction, we conducted an ablation study on the TCGA-COADREAD cohort to assess the contribution of hierarchical multi-level features. The full Shazam model integrates low-, middle-, and high-level representations extracted from each teacher model. To isolate the effect of feature depth, we evaluated reduced variants that use only low-level, mid-level, and high-level features or paired combinations of \{low, mid\}, \{low, high\}, or \{mid, high\}. To ensure a controlled comparison, all variants were trained under the same data splits, optimization configuration, and student-model design. Thus, changes in performance directly reflect the contribution of the feature-level combinations being tested. A complete comparison of all feature-level variants is summarized in Supplementary Table~\ref{tab:SurvivalAblation}.

\subsection{Tile-level Ablation Study}

To quantify the contribution of the Mixture of Experts (MoE) module in tile-level classification, we conducted an ablation in which the MoE fusion was removed and the five teacher features were directly stacked without adaptive gating. All other aspects of the training pipeline—including data partitions, optimization scheme, and student model architecture—were kept unchanged, ensuring that any performance difference arises specifically from the removal of the MoE fusion mechanism. By comparing the full MoE-enabled model with its MoE-ablated counterpart on the PCAM and CCRCC datasets, we assess how much adaptive multi-teacher fusion improves tile-level prediction performance. Detailed results for this ablation are provided in Supplementary Table~\ref{tab:TileAblationPCAM} and ~\ref{tab:TileAblationCCRCC}.

%% file: supplement/3-additional.tex
\heading{Computing hardware and software}\\
All experiments were conducted in the Python environment (Python~3.10.0). 
Model training and inference were performed using PyTorch (version~2.8.0, compiled with CUDA~12.8), 
together with TorchVision~0.23.0. Downstream machine learning experiments used 
Scikit-learn~1.7.2 and NumPy~2.2.6. Whole-slide image (WSI) processing relied on the 
OpenSlide-Python interface (version~1.4.2) as well as the CLAM \cite{lu2021data} codebase. Pathology VQA evaluation was conducted using the 
MUMC~\cite{mumc} codebase. All computations were executed on a workstation equipped with 8~$\times$ NVIDIA GeForce RTX~4090 GPUs.

\heading{Data availability}

The HEST-1K dataset used in this study is publicly available through Hugging Face at \url{https://huggingface.co/datasets/MahmoodLab/hest}. Clinical and genomic data for survival analysis experiments can be accessed via the Genomic Data Commons (GDC) Data Portal (\url{https://portal.gdc.cancer.gov/}). The PathVQA dataset used for visual question answering experiments is available at \url{https://github.com/UCSD-AI4H/PathVQA}.

All tile-level classification experiments were conducted using publicly available histopathology datasets, including HunCRC (\url{https://www.cancerimagingarchive.net/collection/hungarian-colorectal-screening/}), UniToPatho (\url{https://github.com/EIDOSLAB/UNITOPATHO}), PCAM (\url{https://github.com/basveeling/pcam}), ESCA (\url{https://zenodo.org/records/7548828}), PanCancer-TIL (\url{https://zenodo.org/records/6604094}), CRC-100K (\url{https://zenodo.org/records/1214456}), CCRCC-TCGA-HEL (\url{https://zenodo.org/records/7898308}), BACH (\url{https://zenodo.org/records/3632035}), CRC-MSI (\url{https://zenodo.org/records/3832231}), PanCancer-TCGA (\url{https://zenodo.org/records/5889558}), and the Chaoyang dataset (\url{https://github.com/bupt-ai-cz/HSA-NRL}).

Processed data and model outputs generated during this study are available from the corresponding author upon reasonable request.

\heading{Code availability}\\

The code of the Shazam is available on GitHub {\url{https://github.com/Tuner12/Shazam}}

\heading{Author contributions}\\

Conceptualization: W.L., Y.T., A.L., H.C., X.Z., and S.Z. Methodology: W.L., Y.T., A.L., X.Z., and S.Z. Investigation: W.L., Y.T., A.L., H.C., H.T., R.L., F.Y. Visualization: W.L., Y.T., A.L. Funding acquisition: X.Z., S.Z. Supervision: X.Z., S.Z. Writing: W.L., Y.T., A.L.

\heading{Acknowledgements}\\

\clearpage

\heading{Supplementary Files}\\
\renewcommand{\figurename}{Supplementary Figure} 
\renewcommand{\thefigure}{\arabic{figure}}
\setcounter{figure}{0}

\includegraphics[width=1.0\linewidth]{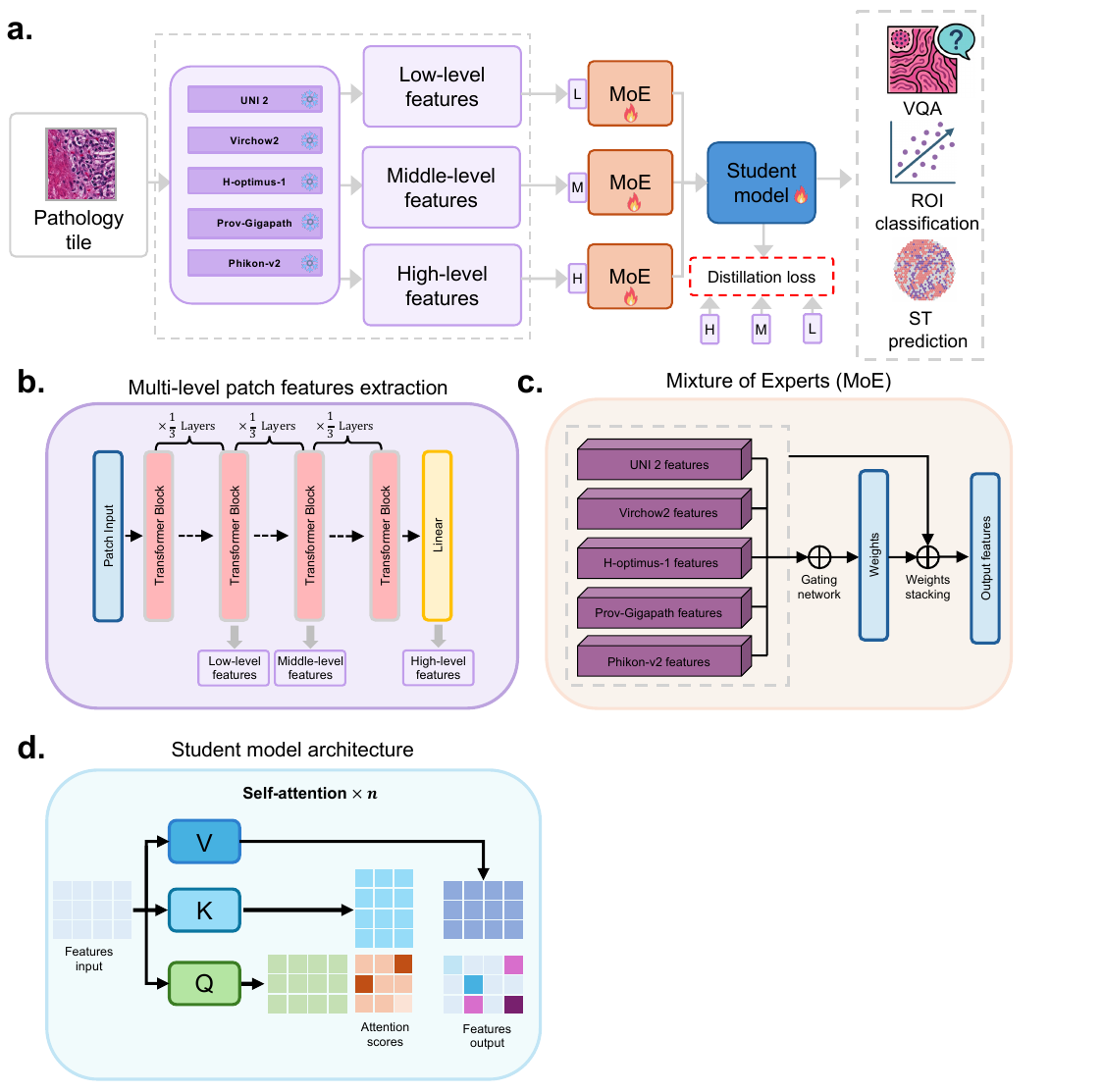}

\captionof{figure}{
Shazam architecture for tile-level pathology analysis.
\textbf{a}, Overview of the Shazam architecture for tile-level classification, illustrating how multi-scale features from multiple pathology foundation models are integrated.
\textbf{b}, Multi-level feature extraction illustrating how low-, middle-, and high-level features are obtained by sampling the transformer at the one-third, two-thirds, and final depth positions.
\textbf{c}, Mixture of Experts (MoE) module used to fuse features from all foundation models at each feature level through an adaptive gating mechanism.
\textbf{d}, Student model structure, which applies a four-layer self-attention module to the fused representations to produce the final tile-level embeddings for downstream prediction tasks.
}

\label{fig:hest-gene-1}

\includegraphics[width=1.0\linewidth]{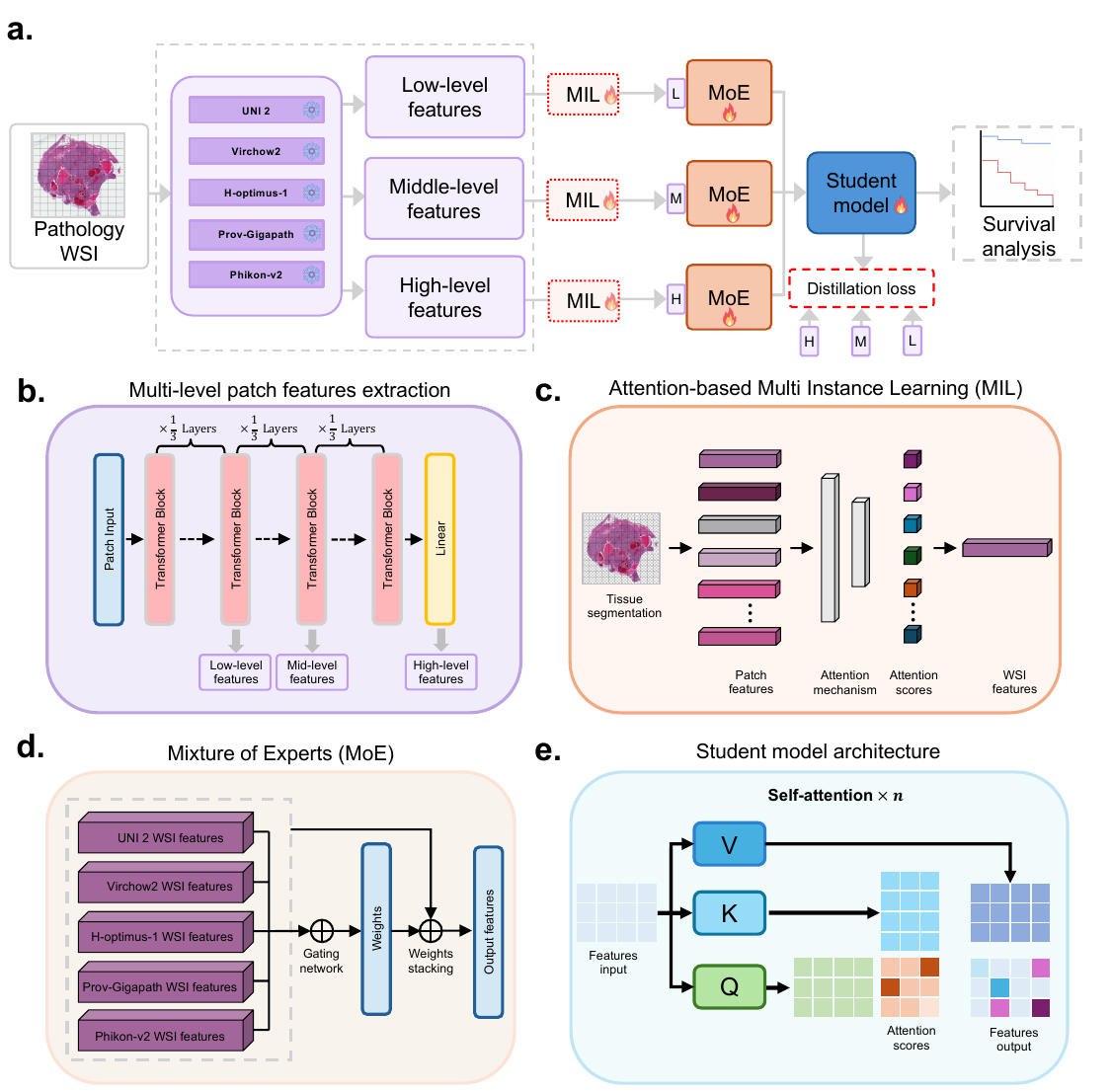}
\captionof{figure}{
Shazam architecture for WSI-level pathology analysis.
\textbf{a}, Overview of the Shazam architecture for WSI-level prediction, illustrating how multi-scale features extracted from pathology foundation models are aggregated and integrated.
\textbf{b}, Multi-level tile feature extraction illustrating how low-, middle-, and high-level features are obtained by sampling the transformer at the one-third, two-thirds, and final depth positions.
\textbf{c}, Attention-based Multiple Instance Learning (ABMIL) used to aggregate tile embeddings into a slide-level representation through an attention mechanism.
\textbf{d}, Mixture of Experts (MoE) module used to fuse slide-level features from all foundation models at each feature level through an adaptive gating mechanism.
\textbf{e}, Student model structure, which applies a four-layer self-attention module to the fused WSI-level representations to produce the final slide-level embeddings for downstream survival prediction.
}

\label{fig:hest-gene-1}

\renewcommand{\tablename}{Supplementary Table}
\renewcommand{\thetable}{\arabic{table}}
\setcounter{table}{0}
\begin{table*}[!ht]
\centering
\caption{\textbf{Cancer-specific gene lists for HEST-1k (Part 1/2).} Gene prioritization was informed by the FDA-recognized content in OncoKB \cite{oncokb_fda_recognition}.}
\label{tab:cancer_gene_lists_csv_1}
\setlength{\tabcolsep}{4pt}
\renewcommand{\arraystretch}{1.15}
\begin{tabularx}{\textwidth}{l >{\raggedright\arraybackslash}X}
\toprule
Cancer & Gene list \\
\midrule
Breast & \scriptsize ALK;\allowbreak APC;\allowbreak AR;\allowbreak ARAF;\allowbreak ARID1A;\allowbreak ATM;\allowbreak BRAF;\allowbreak BRCA1;\allowbreak BRCA2;\allowbreak CCND1;\allowbreak CD274;\allowbreak CDH1;\allowbreak CDK4;\allowbreak CDK6;\allowbreak CDKN2A;\allowbreak CHEK2;\allowbreak CLDN18;\allowbreak CTNNB1;\allowbreak DDR2;\allowbreak DNMT3A;\allowbreak EGFR;\allowbreak EPCAM;\allowbreak ERBB2;\allowbreak ERBB3;\allowbreak ESR1;\allowbreak FBXW7;\allowbreak FGFR1;\allowbreak FGFR2;\allowbreak FGFR3;\allowbreak FGFR4;\allowbreak GNA11;\allowbreak GNAQ;\allowbreak GNAS;\allowbreak HRAS;\allowbreak IDH1;\allowbreak IDH2;\allowbreak JAK1;\allowbreak JAK2;\allowbreak KDR;\allowbreak KEAP1;\allowbreak KIT;\allowbreak KRAS;\allowbreak MAP2K1;\allowbreak MAP2K2;\allowbreak MDM2;\allowbreak MET;\allowbreak MLH1;\allowbreak MSH2;\allowbreak MSH3;\allowbreak MSH6;\allowbreak MTOR;\allowbreak MYC;\allowbreak NF1;\allowbreak NRAS;\allowbreak NRG1;\allowbreak NTRK1;\allowbreak NTRK2;\allowbreak NTRK3;\allowbreak PALB2;\allowbreak PBRM1;\allowbreak PDGFRA;\allowbreak PIK3CA;\allowbreak PIK3R1;\allowbreak PMS2;\allowbreak POLD1;\allowbreak POLE;\allowbreak PTEN;\allowbreak RAF1;\allowbreak RB1;\allowbreak RET;\allowbreak RNF43;\allowbreak ROS1;\allowbreak RSPO2;\allowbreak RSPO3;\allowbreak SDHB;\allowbreak SMAD4;\allowbreak SMARCA4;\allowbreak STK11;\allowbreak TERT;\allowbreak TP53;\allowbreak TSC1;\allowbreak TSC2;\allowbreak VHL; \\
Bowel & \scriptsize ALK;\allowbreak APC;\allowbreak ARID1A;\allowbreak ATM;\allowbreak BRAF;\allowbreak BRCA1;\allowbreak BRCA2;\allowbreak CCND1;\allowbreak CD274;\allowbreak CDH1;\allowbreak CDK4;\allowbreak CDK6;\allowbreak CDKN2A;\allowbreak CHEK2;\allowbreak CTNNB1;\allowbreak DDR2;\allowbreak DNMT3A;\allowbreak EGFR;\allowbreak EPCAM;\allowbreak ERBB2;\allowbreak ERBB3;\allowbreak FBXW7;\allowbreak FGFR1;\allowbreak FGFR2;\allowbreak FGFR3;\allowbreak FGFR4;\allowbreak GNAS;\allowbreak HRAS;\allowbreak IDH1;\allowbreak IDH2;\allowbreak JAK1;\allowbreak JAK2;\allowbreak KDR;\allowbreak KEAP1;\allowbreak KIT;\allowbreak KRAS;\allowbreak MAP2K1;\allowbreak MDM2;\allowbreak MET;\allowbreak MLH1;\allowbreak MSH2;\allowbreak MSH3;\allowbreak MSH6;\allowbreak MTOR;\allowbreak MYC;\allowbreak NF1;\allowbreak NRAS;\allowbreak NRG1;\allowbreak NTRK1;\allowbreak NTRK2;\allowbreak NTRK3;\allowbreak PALB2;\allowbreak PBRM1;\allowbreak PDGFRA;\allowbreak PIK3CA;\allowbreak PIK3R1;\allowbreak PMS2;\allowbreak POLD1;\allowbreak POLE;\allowbreak PTEN;\allowbreak RAF1;\allowbreak RB1;\allowbreak RET;\allowbreak RNF43;\allowbreak ROS1;\allowbreak RSPO2;\allowbreak RSPO3;\allowbreak SDHB;\allowbreak SMAD4;\allowbreak SMARCA4;\allowbreak STK11;\allowbreak TERT;\allowbreak TP53; \\
Kidney & \scriptsize CDKN2A;\allowbreak FGFR2;\allowbreak MED12;\allowbreak PIK3CA;\allowbreak TERT;\allowbreak ALK;\allowbreak CDKN2B;\allowbreak FGFR3;\allowbreak MET;\allowbreak PMS2;\allowbreak TFE3;\allowbreak APC;\allowbreak CREBBP;\allowbreak FH;\allowbreak MLH1;\allowbreak POLE;\allowbreak TFEB;\allowbreak AR;\allowbreak CYP2C8;\allowbreak FLCN;\allowbreak MSH2;\allowbreak PTEN;\allowbreak TP53;\allowbreak ARID1A;\allowbreak DHFR;\allowbreak FLT3;\allowbreak MSH6;\allowbreak RAF1;\allowbreak TSC1;\allowbreak ARID2;\allowbreak DPYD;\allowbreak GNAQ;\allowbreak MTHFR;\allowbreak RB1;\allowbreak TSC2;\allowbreak ATIC;\allowbreak DYNC2H1;\allowbreak GNAS;\allowbreak MTOR;\allowbreak RET;\allowbreak UGT1A1;\allowbreak ATRX;\allowbreak EGFR;\allowbreak GSTP1;\allowbreak MYC;\allowbreak ROS1;\allowbreak VEGFA;\allowbreak BAP1;\allowbreak EMSY;\allowbreak HOXB13;\allowbreak NF1;\allowbreak SDHA;\allowbreak VHL;\allowbreak BMPR1A;\allowbreak EPCAM;\allowbreak HRAS;\allowbreak NFE2L2;\allowbreak SDHB;\allowbreak XPC;\allowbreak BRAF;\allowbreak ERBB2;\allowbreak KDM5C;\allowbreak NOTCH2;\allowbreak SDHC;\allowbreak XRCC1;\allowbreak CBR3;\allowbreak ERCC1;\allowbreak KIT;\allowbreak NRAS;\allowbreak SDHD;\allowbreak CCND1;\allowbreak FGF19;\allowbreak KRAS;\allowbreak NTRK1;\allowbreak SETD2;\allowbreak CDA;\allowbreak FGF3;\allowbreak MAP3K1;\allowbreak NTRK2;\allowbreak SMAD4;\allowbreak CDK4;\allowbreak FGF4;\allowbreak MDM2;\allowbreak NTRK3;\allowbreak SMARCA4;\allowbreak CDK6;\allowbreak FGFR1;\allowbreak MDM4;\allowbreak PBRM1;\allowbreak TCEB1; \\
Brain & \scriptsize ABCC2;\allowbreak ABCC4;\allowbreak ABCG2;\allowbreak ABL1;\allowbreak ABRAXAS1;\allowbreak ALK;\allowbreak APC;\allowbreak AR;\allowbreak ARID1B;\allowbreak ASXL1;\allowbreak ATM;\allowbreak ATR;\allowbreak ATRX;\allowbreak AURKA;\allowbreak BAP1;\allowbreak BARD1;\allowbreak BCL11A;\allowbreak BCL11B;\allowbreak BCORL1;\allowbreak BLM;\allowbreak BRAF;\allowbreak BRCA1;\allowbreak BRCA2;\allowbreak BRD3;\allowbreak BRIP1;\allowbreak C8orf34;\allowbreak CASP3;\allowbreak CASP9;\allowbreak CCDC6;\allowbreak CDH1;\allowbreak CDK1;\allowbreak CDK12;\allowbreak CDK7;\allowbreak CDK8;\allowbreak CDK9;\allowbreak CDKN1C;\allowbreak CDKN2A;\allowbreak CDKN2B;\allowbreak CHEK1;\allowbreak CHEK2;\allowbreak CREBBP;\allowbreak CTNNA1;\allowbreak CUL3;\allowbreak CXCL8;\allowbreak CXCR4;\allowbreak CYP19A1;\allowbreak CYP1B1;\allowbreak CYP2B6;\allowbreak CYP2C8;\allowbreak DDIT3;\allowbreak DDR1;\allowbreak DHFR;\allowbreak DPYD;\allowbreak EGFR;\allowbreak EMSY;\allowbreak EP300;\allowbreak EPCAM;\allowbreak EPHA5;\allowbreak EPHA6;\allowbreak EPHA7;\allowbreak ERBB2;\allowbreak ERCC1;\allowbreak ERCC2;\allowbreak ERCC3;\allowbreak ERCC4;\allowbreak ESR1;\allowbreak ESR2;\allowbreak EWSR1;\allowbreak EZH2;\allowbreak FANCA;\allowbreak FANCC;\allowbreak FANCD2;\allowbreak FANCE;\allowbreak FANCF;\allowbreak FANCG;\allowbreak FANCI;\allowbreak FANCL;\allowbreak FANCM;\allowbreak FGFR1;\allowbreak FGFR2;\allowbreak FGFR3;\allowbreak FLT3;\allowbreak GATA1;\allowbreak GATA4;\allowbreak GEN1;\allowbreak GGH;\allowbreak GSTP1;\allowbreak H1-2;\allowbreak H1-5;\allowbreak H3-3A;\allowbreak H3C2;\allowbreak H3C3;\allowbreak HDAC2;\allowbreak HRAS;\allowbreak IDH1;\allowbreak IDH2;\allowbreak IGF2R;\allowbreak KDR;\allowbreak KIT;\allowbreak KMT2A;\allowbreak KMT2B;\allowbreak KMT2C;\allowbreak KMT2D;\allowbreak KRAS;\allowbreak MAP2K4;\allowbreak MAP3K1;\allowbreak MAP4K1;\allowbreak MAP4K4;\allowbreak MAPK14;\allowbreak MET;\allowbreak MGMT;\allowbreak MLH1;\allowbreak MLH3;\allowbreak MMP14;\allowbreak MMP2;\allowbreak MMP9;\allowbreak MRE11;\allowbreak MSH2;\allowbreak MSH6;\allowbreak MTHFR;\allowbreak MTR;\allowbreak MUTYH;\allowbreak MYCN;\allowbreak NBN;\allowbreak NCOR1;\allowbreak NCOR2;\allowbreak NF1;\allowbreak NOTCH3;\allowbreak NOTCH4;\allowbreak NQO1;\allowbreak NRAS;\allowbreak NTRK1;\allowbreak NTRK2;\allowbreak NTRK3;\allowbreak PALB2;\allowbreak PARP1;\allowbreak PARP2;\allowbreak PARP3;\allowbreak PARP4;\allowbreak PDGFRA;\allowbreak PDGFRB;\allowbreak PIK3C3;\allowbreak PIK3CA;\allowbreak PIK3CD;\allowbreak PML;\allowbreak PMS1;\allowbreak PMS2;\allowbreak POLD1;\allowbreak POLD2;\allowbreak POLE;\allowbreak POLE2;\allowbreak PPM1D;\allowbreak PPP2R2A;\allowbreak PRKDC;\allowbreak PTEN;\allowbreak RAD50;\allowbreak RAD51;\allowbreak RAD51B;\allowbreak RAD51C;\allowbreak RAD51D;\allowbreak RAD52;\allowbreak RAD54B;\allowbreak RAD54L;\allowbreak RB1;\allowbreak RECQL4;\allowbreak RET;\allowbreak ROS1;\allowbreak RPA1;\allowbreak RRM1;\allowbreak SEMA3C;\allowbreak SLC19A1;\allowbreak SLC22A2;\allowbreak SLC28A1;\allowbreak SLCO1B1;\allowbreak SLX4;\allowbreak SMARCB1;\allowbreak SOD2;\allowbreak STAT3;\allowbreak STAT4;\allowbreak STK11;\allowbreak TERT;\allowbreak TGFBR1;\allowbreak TGFBR2;\allowbreak TNFRSF14;\allowbreak TNFRSF17;\allowbreak TP53;\allowbreak TPMT;\allowbreak TSC1;\allowbreak TSC2;\allowbreak UGT1A1;\allowbreak UMPS;\allowbreak WRN;\allowbreak XPC;\allowbreak XRCC1;\allowbreak XRCC2;\allowbreak YAP1;\allowbreak ZBTB16;\allowbreak ZFTA;\allowbreak ZNF217; \\
\bottomrule
\end{tabularx}
\end{table*}

\begin{table*}[!ht]
\centering
\caption{\textbf{Cancer-specific gene lists for HEST-1k (Part 2/2).} Gene prioritization was informed by the FDA-recognized content in OncoKB \cite{oncokb_fda_recognition}.}
\label{tab:cancer_gene_lists_csv_2}
\setlength{\tabcolsep}{4pt}
\renewcommand{\arraystretch}{1.15}
\begin{tabularx}{\textwidth}{l >{\raggedright\arraybackslash}X}
\toprule
Cancer & Gene list \\
\midrule
Skin & \scriptsize CDK6;\allowbreak FANCE;\allowbreak JAK3;\allowbreak NRG1;\allowbreak RUNX1;\allowbreak ACVR1;\allowbreak CDK8;\allowbreak FANCF;\allowbreak JUN;\allowbreak NSD1;\allowbreak RYBP;\allowbreak ACVR1B;\allowbreak CDKN1A;\allowbreak FANCG;\allowbreak KDM5A;\allowbreak NTRK1;\allowbreak SDC4;\allowbreak AKT1;\allowbreak CDKN1B;\allowbreak FANCL;\allowbreak KDM5C;\allowbreak NTRK2;\allowbreak SDHA;\allowbreak AKT2;\allowbreak CDKN2A;\allowbreak FAT1;\allowbreak KDM6A;\allowbreak NTRK3;\allowbreak SDHAF2;\allowbreak AKT3;\allowbreak CDKN2B;\allowbreak FBXW7;\allowbreak KDR;\allowbreak NUTM1;\allowbreak SDHB;\allowbreak ALK;\allowbreak CDKN2C;\allowbreak FGF10;\allowbreak KEAP1;\allowbreak PAK7;\allowbreak SDHC;\allowbreak ALOX12B;\allowbreak CEBPA;\allowbreak FGF14;\allowbreak KEL;\allowbreak PALB2;\allowbreak SDHD;\allowbreak ANKRD11;\allowbreak CHD4;\allowbreak FGF19;\allowbreak KIF5B;\allowbreak PARK2;\allowbreak SETD2;\allowbreak APC;\allowbreak CHEK1;\allowbreak FGF23;\allowbreak KIT;\allowbreak PARP1;\allowbreak SF3B1;\allowbreak AR;\allowbreak CHEK2;\allowbreak FGF3;\allowbreak KLF4;\allowbreak PAX5;\allowbreak SGK1;\allowbreak ARAF;\allowbreak CIC;\allowbreak FGF4;\allowbreak KLHL6;\allowbreak PBRM1;\allowbreak SH2B3;\allowbreak ARFRP1;\allowbreak CREBBP;\allowbreak FGF6;\allowbreak KMT2A;\allowbreak PDCD1;\allowbreak SLC34A2;\allowbreak ARID1A;\allowbreak CRKL;\allowbreak FGFR1;\allowbreak KMT2B;\allowbreak PDCD1LG2;\allowbreak SMAD2;\allowbreak ARID1B;\allowbreak CRLF2;\allowbreak FGFR2;\allowbreak KMT2C;\allowbreak PDGFRA;\allowbreak SMAD3;\allowbreak ARID2;\allowbreak CSF1R;\allowbreak FGFR3;\allowbreak KMT2D;\allowbreak PDGFRB;\allowbreak SMAD4;\allowbreak ARID5B;\allowbreak CSF3R;\allowbreak FGFR4;\allowbreak KRAS;\allowbreak PDK1;\allowbreak SMARCA4;\allowbreak ASXL1;\allowbreak CTCF;\allowbreak FH;\allowbreak LATS1;\allowbreak PIK3C2B;\allowbreak SMARCB1;\allowbreak ASXL2;\allowbreak CTLA4;\allowbreak FLCN;\allowbreak LATS2;\allowbreak PIK3C2G;\allowbreak SMO;\allowbreak ATIC;\allowbreak CTNNA1;\allowbreak FLI1;\allowbreak LMO1;\allowbreak PIK3CA;\allowbreak SNCAIP;\allowbreak ATM;\allowbreak CTNNB1;\allowbreak FLT1;\allowbreak LRP1B;\allowbreak PIK3CB;\allowbreak SOCS1;\allowbreak ATR;\allowbreak CUL3;\allowbreak FLT3;\allowbreak LYN;\allowbreak PIK3CD;\allowbreak SOX17;\allowbreak ATRX;\allowbreak CUX1;\allowbreak FLT4;\allowbreak MAP2K1;\allowbreak PIK3R1;\allowbreak SOX2;\allowbreak AURKA;\allowbreak CXCR4;\allowbreak FOXA1;\allowbreak MAP2K2;\allowbreak PIK3R2;\allowbreak SOX9;\allowbreak AURKB;\allowbreak CYP2C8;\allowbreak FOXL2;\allowbreak MAP2K4;\allowbreak PIM1;\allowbreak SPEN;\allowbreak AXIN1;\allowbreak CYP2D6;\allowbreak FOXO1;\allowbreak MAP3K1;\allowbreak PMS1;\allowbreak SPOP;\allowbreak AXIN2;\allowbreak DAXX;\allowbreak FOXP1;\allowbreak MAP3K13;\allowbreak PMS2;\allowbreak SRC;\allowbreak AXL;\allowbreak DCUN1D1;\allowbreak FUBP1;\allowbreak MAPK1;\allowbreak POLD1;\allowbreak STAG2;\allowbreak B2M;\allowbreak DDR2;\allowbreak GABRA6;\allowbreak MAPK3;\allowbreak POLE;\allowbreak STAT3;\allowbreak BAP1;\allowbreak DHFR;\allowbreak GATA3;\allowbreak MAX;\allowbreak PPARG;\allowbreak STAT5B;\allowbreak BARD1;\allowbreak DICER1;\allowbreak GATA4;\allowbreak MCL1;\allowbreak PPM1D;\allowbreak STK11;\allowbreak BBC3;\allowbreak DNMT1;\allowbreak GATA6;\allowbreak MDM2;\allowbreak PPP2R1A;\allowbreak SUFU;\allowbreak BCL10;\allowbreak DNMT3A;\allowbreak GID4;\allowbreak MDM4;\allowbreak PPP2R2A;\allowbreak SYK;\allowbreak BCL2;\allowbreak DNMT3B;\allowbreak GLI1;\allowbreak MED12;\allowbreak PPP6C;\allowbreak TBX3;\allowbreak BCL2L1;\allowbreak DOT1L;\allowbreak GNA11;\allowbreak MEF2B;\allowbreak PRDM1;\allowbreak TCF3;\allowbreak BCL2L11;\allowbreak DPYD;\allowbreak GNA13;\allowbreak MEN1;\allowbreak PREX2;\allowbreak TCF7L2;\allowbreak BCL2L2;\allowbreak DYNC2H1;\allowbreak GNAQ;\allowbreak MET;\allowbreak PRKAR1A;\allowbreak TEK;\allowbreak BCL6;\allowbreak EED;\allowbreak GNAS;\allowbreak MGA;\allowbreak PRKCI;\allowbreak TERT;\allowbreak BCOR;\allowbreak EGFR;\allowbreak GPS2;\allowbreak MITF;\allowbreak PTCH1;\allowbreak TET1;\allowbreak BCORL1;\allowbreak EIF1AX;\allowbreak GRIN2A;\allowbreak MLH1;\allowbreak PTEN;\allowbreak TET2;\allowbreak BCR;\allowbreak EML4;\allowbreak GRM3;\allowbreak MPL;\allowbreak PTPN11;\allowbreak TGFBR2;\allowbreak BLM;\allowbreak EMSY;\allowbreak GSK3B;\allowbreak MRE11A;\allowbreak PTPRD;\allowbreak TMPRSS2;\allowbreak BMPR1A;\allowbreak EP300;\allowbreak GSTM1;\allowbreak MSH2;\allowbreak PTPRS;\allowbreak TNFAIP3;\allowbreak BRAF;\allowbreak EPCAM;\allowbreak GSTP1;\allowbreak MSH3;\allowbreak PTPRT;\allowbreak TNFRSF14;\allowbreak BRCA1;\allowbreak EPHA3;\allowbreak GSTT1;\allowbreak MSH6;\allowbreak QKI;\allowbreak TOP1;\allowbreak BRCA2;\allowbreak EPHA5;\allowbreak H3F3A;\allowbreak MST1;\allowbreak RAC1;\allowbreak TP53;\allowbreak BRD4;\allowbreak EPHA7;\allowbreak HDAC1;\allowbreak MST1R;\allowbreak RAD21;\allowbreak TP63;\allowbreak BRIP1;\allowbreak EPHB1;\allowbreak HGF;\allowbreak MTHFR;\allowbreak RAD50;\allowbreak TRAF7;\allowbreak BTG1;\allowbreak ERBB2;\allowbreak HIST1H3B;\allowbreak MTOR;\allowbreak RAD51;\allowbreak TSC1;\allowbreak BTK;\allowbreak ERBB3;\allowbreak HLA-A;\allowbreak MUTYH;\allowbreak RAD51B;\allowbreak TSC2;\allowbreak CALR;\allowbreak ERBB4;\allowbreak HLA-B;\allowbreak MYB;\allowbreak RAD51C;\allowbreak TSHR;\allowbreak CARD11;\allowbreak ERCC1;\allowbreak HNF1A;\allowbreak MYC;\allowbreak RAD51D;\allowbreak U2AF1;\allowbreak CASP8;\allowbreak ERCC2;\allowbreak HOXB13;\allowbreak MYCN;\allowbreak RAD52;\allowbreak UGT1A1;\allowbreak CBFB;\allowbreak ERCC3;\allowbreak HRAS;\allowbreak MYD88;\allowbreak RAD54L;\allowbreak VEGFA;\allowbreak CBL;\allowbreak ERCC4;\allowbreak HSD3B1;\allowbreak MYOD1;\allowbreak RAF1;\allowbreak VHL;\allowbreak CBR3;\allowbreak ERCC5;\allowbreak ID3;\allowbreak NBN;\allowbreak RANBP2;\allowbreak VTCN1;\allowbreak CCND1;\allowbreak ERG;\allowbreak IDH1;\allowbreak NCOR1;\allowbreak RARA;\allowbreak WHSC1;\allowbreak CCND2;\allowbreak ERRFI1;\allowbreak IDH2;\allowbreak NF1;\allowbreak RASA1;\allowbreak WHSC1L1;\allowbreak CCND3;\allowbreak ESR1;\allowbreak IGF1;\allowbreak NF2;\allowbreak RB1;\allowbreak WT1;\allowbreak CCNE1;\allowbreak ETV1;\allowbreak IGF1R;\allowbreak NFE2L2;\allowbreak RBM10;\allowbreak XPC;\allowbreak CD274;\allowbreak ETV4;\allowbreak IGF2;\allowbreak NFKBIA;\allowbreak RECQL4;\allowbreak XPO1;\allowbreak CD276;\allowbreak ETV5;\allowbreak IKBKE;\allowbreak NKX3-1;\allowbreak REL;\allowbreak XRCC1;\allowbreak CD74;\allowbreak ETV6;\allowbreak IKZF1;\allowbreak NOTCH;\allowbreak RET;\allowbreak XRCC2;\allowbreak CD79A;\allowbreak EWSR1;\allowbreak INHA;\allowbreak NOTCH1;\allowbreak RHOA;\allowbreak ZNF217;\allowbreak CD79B;\allowbreak EZH2;\allowbreak INPP4B;\allowbreak NOTCH2;\allowbreak RICTOR;\allowbreak ZNF703;\allowbreak CDA;\allowbreak EZR;\allowbreak IRF2;\allowbreak NOTCH3;\allowbreak RIT1;\allowbreak CDC73;\allowbreak FAM175A;\allowbreak IRF4;\allowbreak NOTCH4;\allowbreak RNF43;\allowbreak CDH1;\allowbreak FAM46C;\allowbreak IRS2;\allowbreak NPM1;\allowbreak ROS1;\allowbreak CDK12;\allowbreak FANCA;\allowbreak JAK1;\allowbreak NQO1;\allowbreak RPTOR;\allowbreak CDK4;\allowbreak FANCC;\allowbreak JAK2;\allowbreak NRAS;\allowbreak RSPO2; \\
Liver & \scriptsize CCND1;\allowbreak FGF19;\allowbreak MDM2;\allowbreak PALB2;\allowbreak TP53;\allowbreak ALK;\allowbreak CDA;\allowbreak FGF3;\allowbreak MET;\allowbreak PBRM1;\allowbreak TSC1;\allowbreak APC;\allowbreak CDKN2A;\allowbreak FGF4;\allowbreak MLH1;\allowbreak PIK3CA;\allowbreak TSC2;\allowbreak ARID1A;\allowbreak CDKN2B;\allowbreak FGFR1;\allowbreak MSH2;\allowbreak PMS2;\allowbreak UGT1A1;\allowbreak ARID2;\allowbreak CTNNB1;\allowbreak FGFR2;\allowbreak MSH6;\allowbreak POLD1;\allowbreak VEGFA;\allowbreak ATIC;\allowbreak CYP2C8;\allowbreak FGFR3;\allowbreak MTHFR;\allowbreak POLE;\allowbreak XPC;\allowbreak ATM;\allowbreak DHFR;\allowbreak GSTP1;\allowbreak MTOR;\allowbreak PTEN;\allowbreak XRCC1;\allowbreak ATR;\allowbreak DPYD;\allowbreak IDH1;\allowbreak MYC;\allowbreak RAD51D;\allowbreak BAP1;\allowbreak DYNC2H1;\allowbreak IDH2;\allowbreak NF1;\allowbreak RB1;\allowbreak BRAF;\allowbreak EGFR;\allowbreak JAK1;\allowbreak NRAS;\allowbreak RET;\allowbreak BRCA1;\allowbreak EPCAM;\allowbreak KEAP1;\allowbreak NTRK1;\allowbreak ROS1;\allowbreak BRCA2;\allowbreak ERBB2;\allowbreak KRAS;\allowbreak NTRK2;\allowbreak SMAD4;\allowbreak CBR3;\allowbreak ERCC1;\allowbreak MAP3K1;\allowbreak NTRK3;\allowbreak TERT; \\
Prostate & \scriptsize BRIP1;\allowbreak CYP2C8;\allowbreak HOXB13;\allowbreak NTRK1;\allowbreak RAD54L;\allowbreak ALK;\allowbreak CBR3;\allowbreak DHFR;\allowbreak HRAS;\allowbreak NTRK2;\allowbreak RAF1;\allowbreak APC;\allowbreak CCND1;\allowbreak DPYD;\allowbreak IDH1;\allowbreak NTRK3;\allowbreak RB1;\allowbreak AR;\allowbreak CDA;\allowbreak DYNC2H1;\allowbreak KRAS;\allowbreak PALB2;\allowbreak RET;\allowbreak ATIC;\allowbreak CDH1;\allowbreak EGFR;\allowbreak MAP3K1;\allowbreak PIK3CA;\allowbreak SPOP;\allowbreak ATM;\allowbreak CDK12;\allowbreak EPCAM;\allowbreak MED12;\allowbreak PMS2;\allowbreak TMPRSS2;\allowbreak ATR;\allowbreak CDK4;\allowbreak ERBB2;\allowbreak MLH1;\allowbreak POLE;\allowbreak TP53;\allowbreak ATRX;\allowbreak CDK6;\allowbreak ERCC1;\allowbreak MRE11A;\allowbreak PPP2R2A;\allowbreak UGT1A1;\allowbreak BAP1;\allowbreak CDKN2A;\allowbreak FAM175A;\allowbreak MSH2;\allowbreak PTEN;\allowbreak XPC;\allowbreak BARD1;\allowbreak CDKN2B;\allowbreak FANCA;\allowbreak MSH6;\allowbreak RAD50;\allowbreak XRCC1;\allowbreak BRAF;\allowbreak CHEK1;\allowbreak FANCL;\allowbreak MTHFR;\allowbreak RAD51B;\allowbreak BRCA1;\allowbreak CHEK2;\allowbreak FOXA1;\allowbreak MYC;\allowbreak RAD51C;\allowbreak BRCA2;\allowbreak CTNNB1;\allowbreak GSTP1;\allowbreak NBN;\allowbreak RAD51D; \\
Bladder & \scriptsize CDA;\allowbreak ERBB2;\allowbreak MAP3K1;\allowbreak NTRK2;\allowbreak RET;\allowbreak APC;\allowbreak CDK12;\allowbreak ERCC1;\allowbreak MDM2;\allowbreak NTRK3;\allowbreak ROS1;\allowbreak ARID1A;\allowbreak CDK4;\allowbreak FANCA;\allowbreak MET;\allowbreak PALB2;\allowbreak SDHA;\allowbreak ARID2;\allowbreak CDK6;\allowbreak FANCL;\allowbreak MLH1;\allowbreak PBRM1;\allowbreak SETD2;\allowbreak ATIC;\allowbreak CDKN2A;\allowbreak FBXW7;\allowbreak MRE11A;\allowbreak PIK3CA;\allowbreak SMARCA4;\allowbreak ATM;\allowbreak CDKN2B;\allowbreak FGF19;\allowbreak MSH2;\allowbreak PMS2;\allowbreak SOX2;\allowbreak ATR;\allowbreak CHEK1;\allowbreak FGF3;\allowbreak MSH6;\allowbreak POLE;\allowbreak STK11;\allowbreak ATRX;\allowbreak CHEK2;\allowbreak FGF4;\allowbreak MTHFR;\allowbreak PPP2R2A;\allowbreak TERT;\allowbreak BAP1;\allowbreak CREBBP;\allowbreak FGFR1;\allowbreak MTOR;\allowbreak PTEN;\allowbreak TP53;\allowbreak BARD1;\allowbreak CTNNB1;\allowbreak FGFR2;\allowbreak MUTYH;\allowbreak RAD50;\allowbreak TSC1;\allowbreak BRAF;\allowbreak CYP2C8;\allowbreak FGFR3;\allowbreak MYC;\allowbreak RAD51B;\allowbreak TSC2;\allowbreak BRCA1;\allowbreak DHFR;\allowbreak GATA3;\allowbreak NBN;\allowbreak RAD51C;\allowbreak UGT1A1;\allowbreak BRCA2;\allowbreak DPYD;\allowbreak GSTP1;\allowbreak NF1;\allowbreak RAD51D;\allowbreak XPC;\allowbreak BRIP1;\allowbreak DYNC2H1;\allowbreak HRAS;\allowbreak NOTCH2;\allowbreak RAD54L;\allowbreak XRCC1;\allowbreak CBR3;\allowbreak EGFR;\allowbreak KRAS;\allowbreak NRAS;\allowbreak RAF1;\allowbreak CCND1;\allowbreak EPCAM;\allowbreak MAP2K1;\allowbreak NTRK1;\allowbreak RB1; \\
\bottomrule
\end{tabularx}
\end{table*}

\begin{table*}[!ht]
\centering
\caption{\textbf{ST prediction performance of different foundation models on HEST-1k dataset (Part 1/2).} The 95\% CI is included in parentheses. Best model is \textbf{bolded}, second-best is \underline{underlined}.}
\begin{tabular}{l l c r r}
\toprule
Model & Organ & PCC & Num & Test patient num \\
\midrule
UNI 2        & breast   & \underline{0.160 (0.088--0.231)} & 125 & 36 \\
Virchow2     & breast   & 0.135 (0.065--0.205)             & 125 & 36 \\
H-optimus-1  & breast   & 0.145 (0.097--0.194)             & 125 & 36 \\
Prov-Gigapath& breast   & 0.121 (0.056--0.187)             & 125 & 36 \\
Phikon-v2    & breast   & 0.136 (0.056--0.217)             & 125 & 36 \\
Shazam       & breast   & \textbf{0.176 (0.076--0.276)}    & 125 & 36 \\
\midrule
UNI 2        & prostate & 0.138 (-0.082--0.359)            & 60 & 2 \\
Virchow2     & prostate & 0.138 (-0.177--0.453)            & 60 & 2 \\
Prov-Gigapath& prostate & 0.133 (-0.059--0.325)            & 60 & 2 \\
Phikon-v2    & prostate & 0.141 (-0.281--0.562)            & 60 & 2 \\
H-optimus-1  & prostate & \underline{0.156 (-0.223--0.535)}& 60 & 2 \\
Shazam       & prostate & \textbf{0.185 (-0.181--0.550)}   & 60 & 2 \\
\midrule
UNI 2        & skin     & 0.103 (0.048--0.158)             & 27 & 7 \\
Virchow2     & skin     & 0.108 (0.048--0.167)             & 27 & 7 \\
Prov-Gigapath& skin     & 0.099 (0.035--0.162)             & 27 & 7 \\
Phikon-v2    & skin     & 0.100 (0.046--0.154)             & 27 & 7 \\
H-optimus-1  & skin     & \underline{0.108 (0.053--0.163)} & 27 & 7 \\
Shazam       & skin     & \textbf{0.153 (0.072--0.234)}    & 27 & 7 \\
\midrule
UNI 2        & bowel    & \underline{0.193 (0.126--0.259)} & 64 & 6 \\
Virchow2     & bowel    & 0.186 (0.137--0.236)             & 64 & 6 \\
H-optimus-1  & bowel    & 0.165 (0.106--0.225)             & 64 & 6 \\
Prov-Gigapath& bowel    & 0.132 (0.008--0.256)             & 64 & 6 \\
Phikon-v2    & bowel    & 0.156 (0.075--0.237)             & 64 & 6 \\
Shazam       & bowel    & \textbf{0.242 (0.139--0.345)}    & 64 & 6 \\
\bottomrule
\end{tabular}
\label{tab:Table2}
\end{table*}

\begin{table*}[!ht]
\centering
\caption{\textbf{ST prediction performance of different foundation models on HEST-1k dataset (Part 2/2).} The 95\% CI is included in parentheses. Best model is \textbf{bolded}, second-best is \underline{underlined}.}
\begin{tabular}{l l c r r}
\toprule
Model & Organ & PCC & Num & Test patient num \\
\midrule
UNI 2        & liver    & 0.050 (0.048--0.051)              & 13 & 2 \\
Virchow2     & liver    & \underline{0.070 (-0.121--0.262)} & 13 & 2 \\
Prov-Gigapath& liver    & 0.060 (-0.159--0.278)             & 13 & 2 \\
Phikon-v2    & liver    & 0.059 (0.021--0.097)              & 13 & 2 \\
H-optimus-1  & liver    & 0.057 (-0.037--0.152)             & 13 & 2 \\
Shazam       & liver    & \textbf{0.145 (-0.015--0.304)}    & 13 & 2 \\
\midrule
UNI 2        & kidney   & 0.210 (0.161--0.259)              & 25 & 24 \\
Virchow2     & kidney   & 0.201 (0.156--0.246)              & 25 & 24 \\
H-optimus-1  & kidney   & 0.201 (0.167--0.235)              & 25 & 24 \\
Prov-Gigapath& kidney   & 0.207 (0.184--0.230)              & 25 & 24 \\
Phikon-v2    & kidney   & \underline{0.217 (0.190--0.244)}  & 25 & 24 \\
Shazam       & kidney   & \textbf{0.236 (0.187--0.285)}     & 25 & 24 \\
\midrule
UNI 2        & brain    & 0.086 (0.060--0.112)              & 24 & 11 \\
Virchow2     & brain    & 0.083 (0.067--0.100)              & 24 & 11 \\
Prov-Gigapath& brain    & 0.086 (0.068--0.104)              & 24 & 11 \\
Phikon-v2    & brain    & 0.092 (0.074--0.110)              & 24 & 11 \\
H-optimus-1  & brain    & \underline{0.093 (0.067--0.119)}  & 24 & 11 \\
Shazam       & brain    & \textbf{0.127 (0.097--0.158)}     & 24 & 11 \\
\midrule
UNI 2        & bladder  & 0.049 (-0.003--0.101)             & 6 & 5 \\
Virchow2     & bladder  & \underline{0.061 (0.006--0.115)}  & 6 & 5 \\
H-optimus-1  & bladder  & 0.033 (-0.026--0.092)             & 6 & 5 \\
Prov-Gigapath& bladder  & 0.050 (0.005--0.095)              & 6 & 5 \\
Phikon-v2    & bladder  & 0.047 (-0.017--0.112)             & 6 & 5 \\
Shazam       & bladder  & \textbf{0.107 (0.020--0.194)}     & 6 & 5 \\
\bottomrule
\end{tabular}
\label{tab:Table3}
\end{table*}

\clearpage

\includegraphics[width=1.0\linewidth]{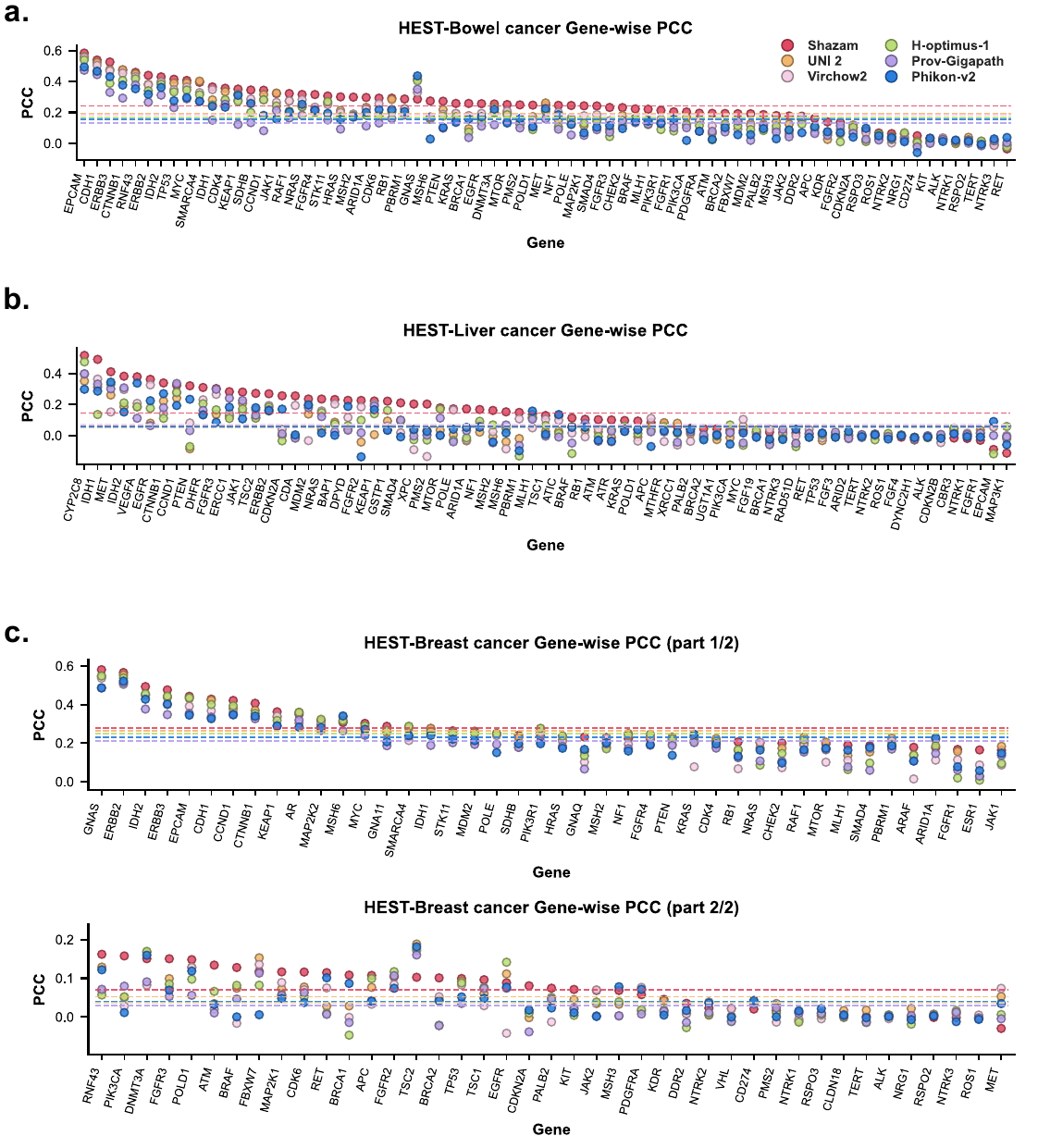}
\captionof{figure}{\textbf{a-c}, Gene-wise PCC for spatial expression prediction in the HEST-Bowel cancer, HEST-Liver cancer, HEST-Breast cancer cohort, comparing Shazam with baseline pathology foundation models.}
\label{fig:hest-gene-1}

\clearpage

\includegraphics[width=1.0\linewidth]{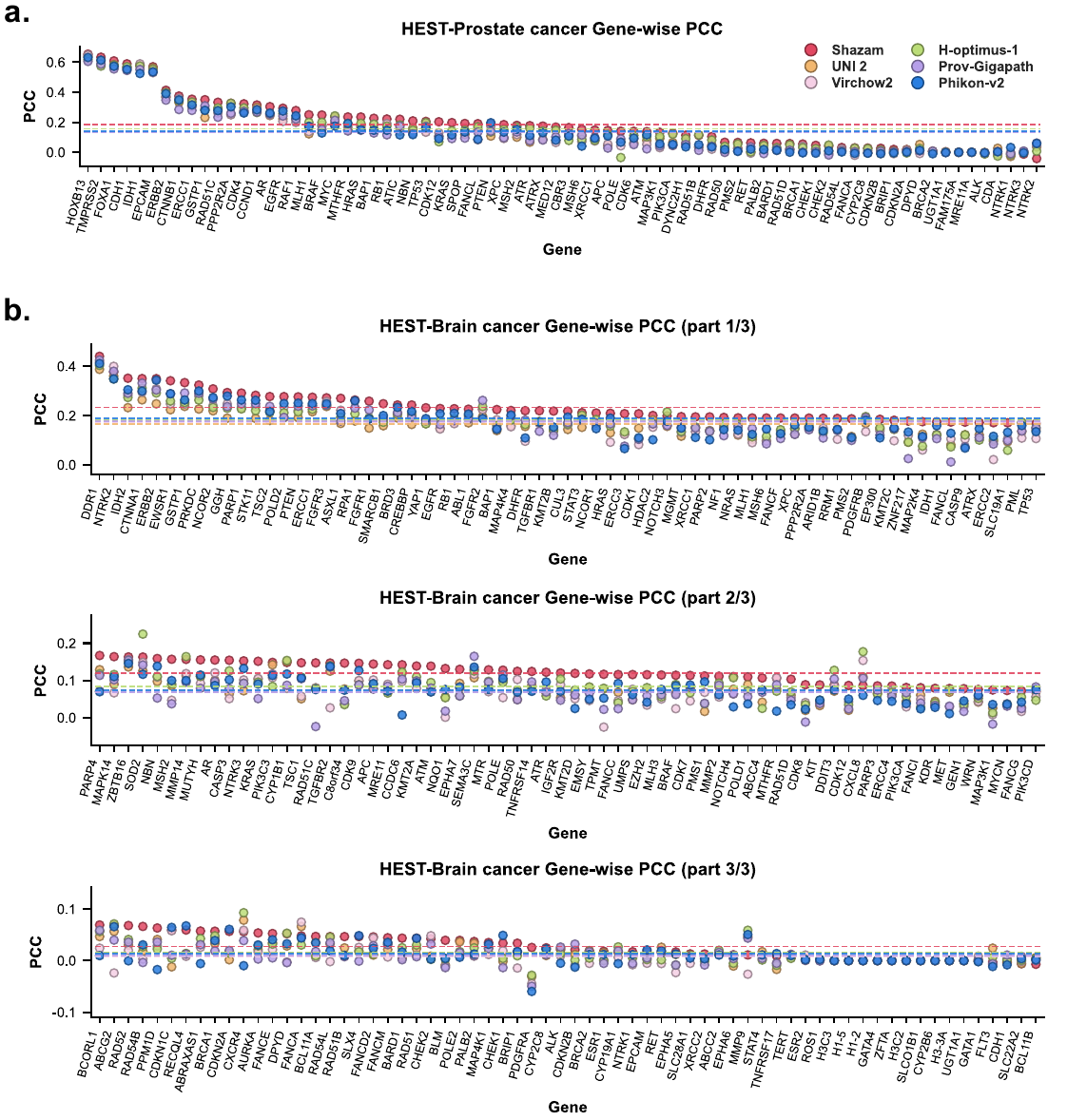}
\captionof{figure}{\textbf{a-b}, Gene-wise PCC for spatial expression prediction in the HEST-Prostate cancer, HEST-Brain cancer cohort, comparing Shazam with baseline pathology foundation models.}
\label{fig:hest-gene-2}

\clearpage

\includegraphics[width=1.0\linewidth]{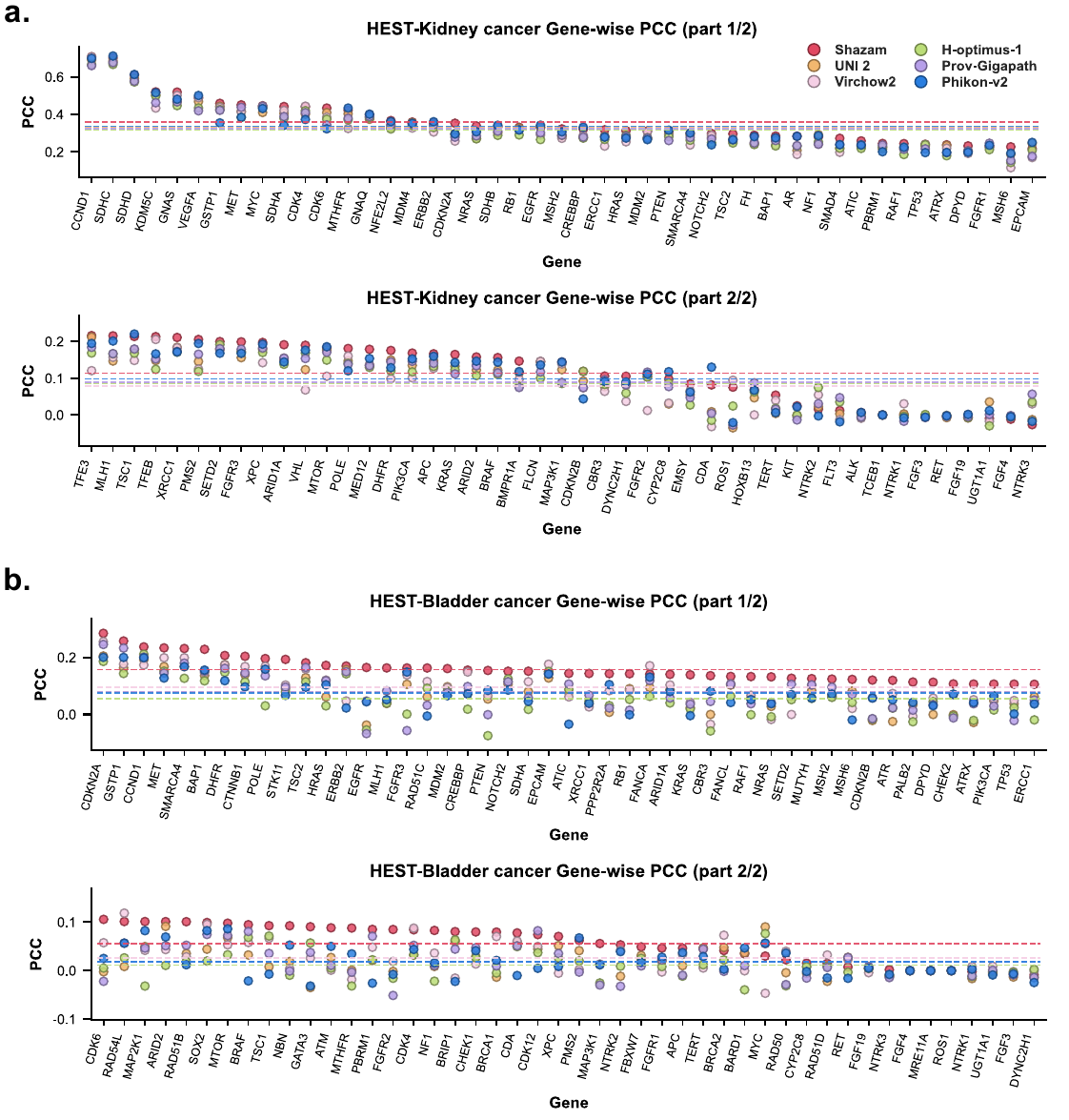}
\captionof{figure}{\textbf{a-b}, Gene-wise PCC for spatial expression prediction in the HEST-Kidney cancer, HEST-Bladder cancer cohort, comparing Shazam with baseline pathology foundation models.}
\label{fig:hest-gene-3}

\clearpage

\includegraphics[width=1.0\linewidth]{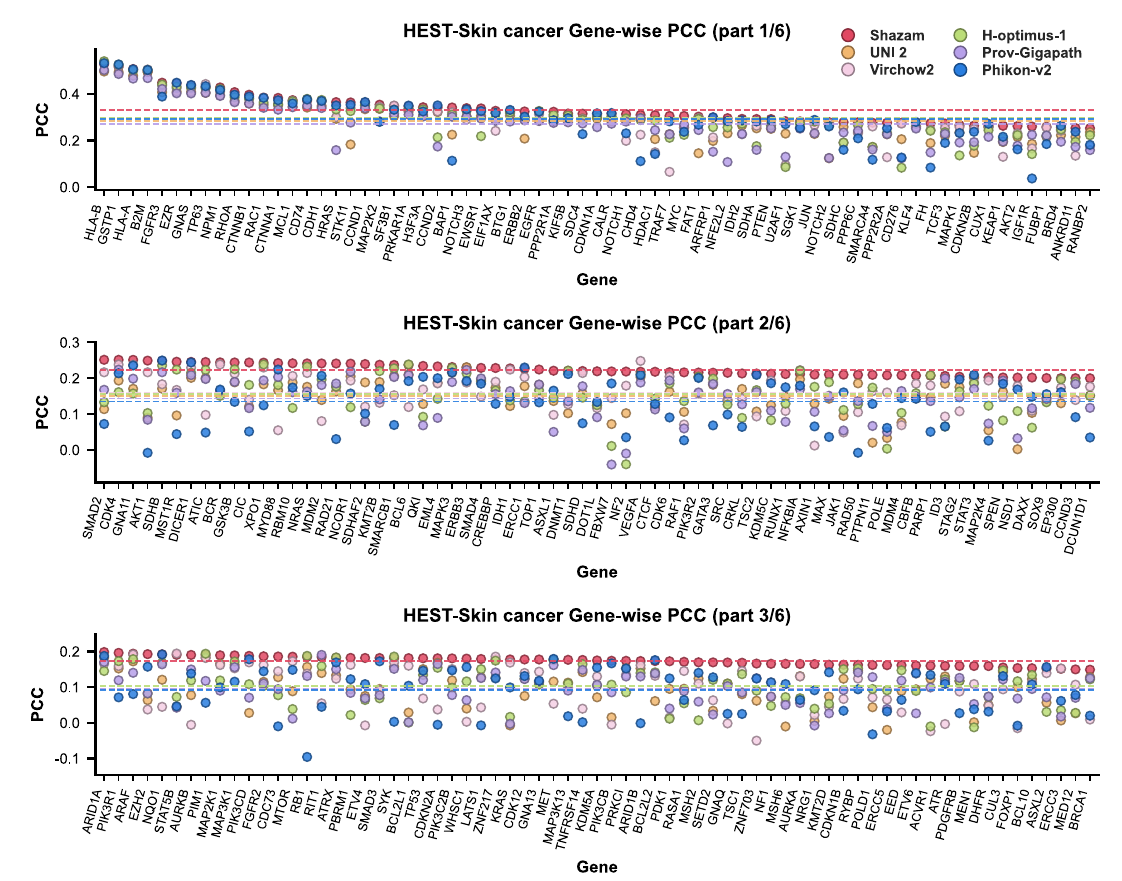}
\captionof{figure}{Gene-wise PCC for spatial expression prediction in the HEST-Skin cancer part 1-3, HEST-Bladder cancer cohort, comparing Shazam with baseline pathology foundation models.}
\label{fig:hest-gene-4}

\clearpage

\includegraphics[width=1.0\linewidth]{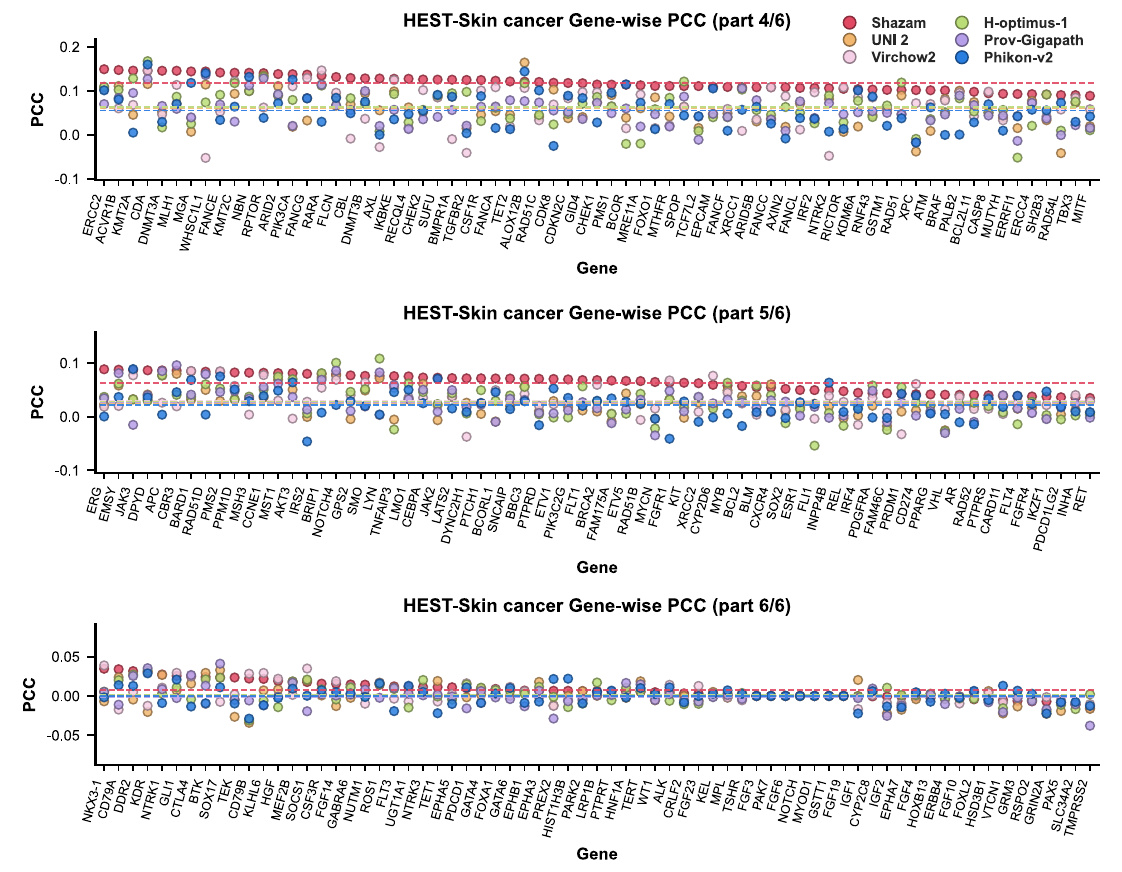}
\captionof{figure}{Gene-wise PCC for spatial expression prediction in the HEST-Skin cancer part 4-6, HEST-Bladder cancer cohort, comparing Shazam with baseline pathology foundation models.}
\label{fig:hest-gene-5}


\begin{table*}[!ht]
\centering
\caption{\textbf{Performance of Survival Analysis on TCGA-BRCA, TCGA-BLCA, and TCGA-CESC Datasets.}
The 95\% CI is included in parentheses. The best and second-best performed models are \textbf{bolded} and \underline{underlined.}}
\label{tab:Table5}
\begin{tabular}{lccc}
\toprule
Model & TCGA-BRCA & TCGA-BLCA & TCGA-CESC \\
\midrule
UNI 2        & 0.627 (0.569--0.685) & 0.598 (0.526--0.669)  & 0.581 (0.406--0.756) \\
Virchow2     & 0.622 (0.557--0.688) & 0.561 (0.485--0.637)  & \underline{0.672 (0.606--0.738)} \\
H-optimus-1  & 0.640 (0.592--0.688) & \underline{0.599 (0.573--0.624)}  & \textbf{0.690 (0.649--0.731)} \\
Prov-Gigapath & 0.631 (0.543--0.718) & 0.539 (0.498--0.580)  & 0.631 (0.540--0.722) \\
Phikon-v2     & \underline{0.662 (0.563--0.760)} & 0.583 (0.506--0.659)  & 0.473 (0.312--0.633) \\
Shazam        & \textbf{0.670 (0.585--0.754)} & \textbf{0.608 (0.555--0.662)}  & 0.666 (0.606--0.727) \\
\bottomrule
\end{tabular}
\end{table*}

\begin{table*}[!ht]
\centering
\caption{\textbf{Performance of Survival Analysis on TCGA-COADREAD, TCGA-GBMLGG, and TCGA-KIRC Datasets.}
The 95\% CI is included in parentheses. The best and second-best performed models are \textbf{bolded} and \underline{underlined.}}
\label{tab:Table6}
\begin{tabular}{lccc}
\toprule
Model & TCGA-COADREAD & TCGA-GBMLGG & TCGA-KIRC \\
\midrule
UNI 2        & 0.596 (0.540--0.652) & 0.756 (0.715--0.797) & 0.687 (0.605--0.769) \\
Virchow2      & 0.625 (0.558--0.691) & 0.756 (0.708--0.805) & 0.694 (0.638--0.750) \\
H-optimus-1   & \underline{0.646 (0.559--0.732)} & \underline{0.767 (0.729--0.804)} & \underline{0.710 (0.640--0.780)} \\
Prov-Gigapath & 0.595 (0.542--0.648) & 0.766 (0.724--0.809) & 0.689 (0.631--0.747) \\
Phikon-v2     & 0.574 (0.516--0.633) & 0.748 (0.715--0.782) & 0.662 (0.584--0.741) \\
Shazam        & \textbf{0.710 (0.655--0.765)} & \textbf{0.768 (0.726--0.811)} & \textbf{0.742 (0.683--0.802)} \\
\bottomrule
\end{tabular}
\end{table*}

\begin{table*}[!ht]
\centering
\caption{\textbf{Performance of Survival Analysis on TCGA-LUAD, TCGA-LUSC, TCGA-SKCM, and TCGA-STAD Datasets.}
The 95\% CI is included in parentheses. The best and second-best performed models are \textbf{bolded} and \underline{underlined.}}
\label{tab:Table7}
\begin{tabular}{lcccc}
\toprule
Model & TCGA-LUAD & TCGA-LUSC & TCGA-SKCM & TCGA-STAD \\
\midrule
UNI 2        & 0.596 (0.493--0.700) & 0.521 (0.433--0.609) & 0.622 (0.589--0.655) & 0.584 (0.512--0.657) \\
Virchow2      & \underline{0.612 (0.520--0.705)} & 0.538 (0.456--0.619) & 0.629 (0.590--0.668) & \underline{0.605 (0.564--0.647)} \\
H-optimus-1   & 0.588 (0.476--0.699) & \underline{0.545 (0.459--0.630)} & \underline{0.637 (0.613--0.662)} & 0.601 (0.554--0.648) \\
Prov-Gigapath & 0.603 (0.474--0.731) & 0.525 (0.450--0.599) & 0.628 (0.604--0.653) & 0.563 (0.517--0.608) \\
Phikon-v2     & 0.571 (0.507--0.635) & 0.498 (0.416--0.579) & 0.581 (0.537--0.625) & 0.591 (0.515--0.668) \\
Shazam        & \textbf{0.628 (0.577--0.679)} & \textbf{0.583 (0.525--0.641)} & \textbf{0.657 (0.613--0.701)} & \textbf{0.626 (0.531--0.721)} \\
\bottomrule
\end{tabular}
\end{table*}

\begin{table*}[!ht]
\centering
\caption{\textbf{Tile-level classification performance of different foundation models on HunCRC dataset.} Non-parametric bootstrapping with 1,000 replicates is employed for statistical analysis. The 95\% CI is included in parentheses. Best performing model is \textbf{bolded} and second-best is \underline{underlined}.}
\label{tab:Table8}
\begin{tabular}{lccc}
\toprule
Model & Balanced ACC & Weighted F1 & Top1-ACC \\
\midrule
UNI 2 & 0.520 (0.492--0.546) & 0.439 (0.421--0.458) & 0.812 (0.807--0.817) \\
Virchow2 & \textbf{0.567 (0.545--0.588)} & \underline{0.443 (0.427--0.460)} & 0.797 (0.792--0.802) \\
H-optimus-1 & 0.458 (0.435--0.483) & 0.414 (0.397--0.430) & 0.829 (0.824--0.833) \\
Prov-Gigapath & 0.472 (0.446--0.499) & 0.428 (0.408--0.448) & \textbf{0.833 (0.828--0.838)} \\
Phikon-v2 & 0.472 (0.446--0.498) & 0.409 (0.393--0.426) & 0.814 (0.809--0.819) \\
Shazam & \underline{0.549 (0.523--0.575)} & \textbf{0.459 (0.442--0.475)} & \underline{0.829 (0.824--0.834)} \\
\bottomrule
\end{tabular}
\end{table*}

\begin{table*}[!ht]
\centering
\caption{\textbf{Tile-level classification performance of different foundation models on UniToPatho dataset.} Non-parametric bootstrapping with 1,000 replicates is employed for statistical analysis. The 95\% CI is included in parentheses. Best performing model is \textbf{bolded} and second-best is \underline{underlined}.}
\label{tab:Table9}
\begin{tabular}{lccc}
\toprule
Model & Balanced ACC & Weighted F1 & Top1-ACC \\
\midrule
UNI 2 & \underline{0.560 (0.539--0.583)} & 0.536 (0.515--0.558) & 0.560 (0.543--0.579) \\
Virchow2 & \textbf{0.606 (0.584--0.628)} & \textbf{0.586 (0.565--0.608)} & \underline{0.610 (0.590--0.630)} \\
H-optimus-1 & 0.465 (0.444--0.486) & 0.423 (0.401--0.443) & 0.505 (0.486--0.525) \\
Prov-Gigapath & 0.498 (0.479--0.518) & 0.508 (0.488--0.529) & 0.595 (0.576--0.616) \\
Phikon-v2 & 0.540 (0.519--0.562) & 0.497 (0.474--0.518) & 0.553 (0.534--0.572) \\
Shazam & 0.532 (0.511--0.553) & \underline{0.548 (0.525--0.570)} & \textbf{0.622 (0.604--0.643)} \\
\bottomrule
\end{tabular}
\end{table*}

\begin{table*}[!ht]
\centering
\caption{\textbf{Tile-level classification performance of different foundation models on PCAM dataset.} Non-parametric bootstrapping with 1,000 replicates is employed for statistical analysis. The 95\% CI is included in parentheses. Best performing model is \textbf{bolded} and second-best is \underline{underlined}.}
\label{tab:Table10}
\begin{tabular}{lccc}
\toprule
Model & Balanced ACC & Weighted F1 & Top1-ACC \\
\midrule
UNI 2 & 0.932 (0.929--0.934) & 0.932 (0.929--0.934) & 0.932 (0.929--0.935) \\
Virchow2 & 0.929 (0.927--0.932) & 0.929 (0.926--0.932) & 0.929 (0.927--0.932) \\
H-optimus-1 & \underline{0.935 (0.932--0.938)} & \underline{0.935 (0.932--0.937)} & \underline{0.935 (0.932--0.938)} \\
Prov-Gigapath & 0.930 (0.927--0.932) & 0.929 (0.927--0.932) & 0.930 (0.927--0.932) \\
Phikon-v2 & 0.886 (0.882--0.889) & 0.885 (0.881--0.888) & 0.886 (0.882--0.889) \\
Shazam & \textbf{0.944 (0.942--0.947)} & \textbf{0.944 (0.941--0.947)} & \textbf{0.944 (0.941--0.947)} \\
\bottomrule
\end{tabular}
\end{table*}

\begin{table*}[!ht]
\centering
\caption{\textbf{Tile-level classification performance of different foundation models on ESCA dataset.} Non-parametric bootstrapping with 1,000 replicates is employed for statistical analysis. The 95\% CI is included in parentheses. Best performing model is \textbf{bolded} and second-best is \underline{underlined}.}
\label{tab:Table11}
\begin{tabular}{lccc}
\toprule
Model & Balanced ACC & Weighted F1 & Top1-ACC \\
\midrule
UNI 2 & 0.808 (0.803--0.813) & 0.832 (0.827--0.836) & 0.945 (0.944--0.946) \\
Virchow2 & \underline{0.848 (0.843--0.853)} & \underline{0.867 (0.863--0.871)} & \textbf{0.956 (0.955--0.957)} \\
H-optimus-1 & 0.812 (0.807--0.817) & 0.837 (0.833--0.842) & 0.951 (0.950--0.952) \\
Prov-Gigapath & 0.782 (0.778--0.787) & 0.787 (0.782--0.791) & 0.921 (0.919--0.922) \\
Phikon-v2 & 0.762 (0.757--0.767) & 0.736 (0.730--0.740) & 0.879 (0.878--0.881) \\
Shazam & \textbf{0.868 (0.863--0.873)} & \textbf{0.881 (0.878--0.885)} & \underline{0.953 (0.952--0.954)} \\
\bottomrule
\end{tabular}
\end{table*}

\begin{table*}[!ht]
\centering
\caption{\textbf{Tile-level classification performance of different foundation models on PanCancer-TIL dataset.} Non-parametric bootstrapping with 1,000 replicates is employed for statistical analysis. The 95\% CI is included in parentheses. Best performing model is \textbf{bolded} and second-best is \underline{underlined}.}
\label{tab:Table12}
\begin{tabular}{lccc}
\toprule
Model & Balanced ACC & Weighted F1 & Top1-ACC \\
\midrule
UNI 2 & \textbf{0.918 (0.915--0.921)} & 0.908 (0.905--0.911) & 0.943 (0.941--0.945) \\
Virchow2 & 0.902 (0.898--0.906) & \underline{0.910 (0.907--0.913)} & \textbf{0.946 (0.945--0.948)} \\
H-optimus-1 & 0.912 (0.908--0.915) & 0.902 (0.899--0.905) & 0.939 (0.937--0.941) \\
Prov-Gigapath & 0.908 (0.904--0.911) & 0.898 (0.895--0.901) & 0.937 (0.935--0.938) \\
Phikon-v2 & 0.913 (0.910--0.916) & 0.909 (0.906--0.912) & 0.944 (0.942--0.946) \\
Shazam & \underline{0.916 (0.912--0.919)} & \textbf{0.912 (0.909--0.915)} & \underline{0.946 (0.944--0.948)} \\
\bottomrule
\end{tabular}
\end{table*}

\begin{table*}[!ht]
\centering
\caption{\textbf{Tile-level classification performance of different foundation models on CRC-100k dataset.} Non-parametric bootstrapping with 1,000 replicates is employed for statistical analysis. The 95\% CI is included in parentheses. Best performing model is \textbf{bolded} and second-best is \underline{underlined}.}
\label{tab:Table13}
\begin{tabular}{lccc}
\toprule
Model & Balanced ACC & Weighted F1 & Top1-ACC \\
\midrule
UNI 2 & \underline{0.953 (0.947--0.958)} & \underline{0.949 (0.943--0.954)} & 0.960 (0.956--0.964) \\
Virchow2 & 0.953 (0.947--0.958) & 0.948 (0.941--0.954) & \underline{0.967 (0.962--0.971)} \\
H-optimus-1 & 0.952 (0.946--0.957) & 0.946 (0.939--0.953) & 0.965 (0.961--0.970) \\
Prov-Gigapath & 0.952 (0.946--0.957) & 0.948 (0.943--0.954) & 0.964 (0.959--0.969) \\
Phikon-v2 & 0.928 (0.922--0.935) & 0.925 (0.918--0.932) & 0.950 (0.945--0.954) \\
Shazam & \textbf{0.960 (0.955--0.966)} & \textbf{0.956 (0.949--0.962)} & \textbf{0.972 (0.968--0.977)} \\
\bottomrule
\end{tabular}
\end{table*}

\begin{table*}[!ht]
\centering
\caption{\textbf{Tile-level classification performance of different foundation models on CCRCC dataset.} Non-parametric bootstrapping with 1,000 replicates is employed for statistical analysis. The 95\% CI is included in parentheses. Best performing model is \textbf{bolded} and second-best is \underline{underlined}.}
\label{tab:Table14}
\begin{tabular}{lccc}
\toprule
Model & Balanced ACC & Weighted F1 & Top1-ACC \\
\midrule
UNI 2 & \underline{0.964 (0.954--0.972)} & 0.960 (0.952--0.967) & \underline{0.974 (0.970--0.979)} \\
Virchow2 & \underline{0.964 (0.955--0.972)} & \underline{0.961 (0.952--0.968)} & 0.972 (0.968--0.976) \\
H-optimus-1 & 0.960 (0.951--0.968) & 0.957 (0.949--0.965) & 0.972 (0.968--0.976) \\
Prov-Gigapath & 0.958 (0.949--0.967) & 0.959 (0.951--0.966) & 0.971 (0.966--0.975) \\
Phikon-v2 & \textbf{0.965 (0.957--0.973)} & 0.959 (0.950--0.967) & 0.973 (0.968--0.977) \\
Shazam & 0.963 (0.953--0.971) & \textbf{0.975 (0.971--0.978)} & \textbf{0.975 (0.971--0.979)} \\
\bottomrule
\end{tabular}
\end{table*}

\begin{table*}[!ht]
\centering
\caption{\textbf{Tile-level classification performance of different foundation models on BACH dataset.} Non-parametric bootstrapping with 1,000 replicates is employed for statistical analysis. The 95\% CI is included in parentheses. Best performing model is \textbf{bolded} and second-best is \underline{underlined}.}
\label{tab:Table15}
\begin{tabular}{lccc}
\toprule
Model & Balanced ACC & Weighted F1 & Top1-ACC \\
\midrule
UNI 2 & \textbf{0.988 (0.958--1.000)} & \textbf{0.987 (0.958--1.000)} & \textbf{0.988 (0.963--1.000)} \\
Virchow2 & \textbf{0.988 (0.958--1.000)} & \textbf{0.987 (0.958--1.000)} & \textbf{0.988 (0.963--1.000)} \\
H-optimus-1 & \underline{0.963 (0.916--1.000)} & \underline{0.962 (0.913--1.000)} & \underline{0.963 (0.912--1.000)} \\
Prov-Gigapath & 0.912 (0.844--0.969) & 0.915 (0.848--0.972) & 0.912 (0.850--0.975) \\
Phikon-v2 & 0.825 (0.738--0.901) & 0.829 (0.741--0.905) & 0.825 (0.738--0.900) \\
Shazam & \textbf{0.988 (0.958--1.000)} & \textbf{0.987 (0.958--1.000)} & \textbf{0.988 (0.963--1.000)} \\
\bottomrule
\end{tabular}
\end{table*}

\begin{table*}[!ht]
\centering
\caption{\textbf{Tile-level classification performance of different foundation models on CRC-MSI dataset.} Non-parametric bootstrapping with 1,000 replicates is employed for statistical analysis. The 95\% CI is included in parentheses. Best performing model is \textbf{bolded} and second-best is \underline{underlined}.}
\label{tab:Table16}
\begin{tabular}{lccc}
\toprule
Model & Balanced ACC & Weighted F1 & Top1-ACC \\
\midrule
UNI 2 & \underline{0.736 (0.729--0.743)} & 0.713 (0.707--0.719) & 0.829 (0.826--0.834) \\
Virchow2 & 0.733 (0.726--0.740) & \underline{0.722 (0.715--0.728)} & \underline{0.842 (0.838--0.846)} \\
H-optimus-1 & 0.721 (0.714--0.728) & 0.700 (0.694--0.706) & 0.822 (0.818--0.826) \\
Prov-Gigapath & 0.730 (0.723--0.737) & 0.693 (0.687--0.699) & 0.806 (0.802--0.810) \\
Phikon-v2 & 0.633 (0.626--0.640) & 0.633 (0.626--0.639) & 0.801 (0.796--0.805) \\
Shazam & \textbf{0.738 (0.731--0.745)} & \textbf{0.726 (0.720--0.732)} & \textbf{0.844 (0.840--0.848)} \\
\bottomrule
\end{tabular}
\end{table*}

\begin{table*}[!ht]
\centering
\caption{\textbf{Tile-level classification performance of different foundation models on PanCancer-TCGA dataset.} Non-parametric bootstrapping with 1,000 replicates is employed for statistical analysis. The 95\% CI is included in parentheses. Best performing model is \textbf{bolded} and second-best is \underline{underlined}.}
\label{tab:Table17}
\begin{tabular}{lccc}
\toprule
Model & Balanced ACC & Weighted F1 & Top1-ACC \\
\midrule
UNI 2 & 0.919 (0.916--0.922) & 0.925 (0.922--0.927) & 0.938 (0.936--0.940) \\
Virchow2 & 0.914 (0.911--0.917) & 0.920 (0.917--0.923) & 0.934 (0.933--0.936) \\
H-optimus-1 & \underline{0.927 (0.924--0.930)} & \textbf{0.932 (0.930--0.935)} & \textbf{0.946 (0.945--0.948)} \\
Prov-Gigapath & 0.909 (0.906--0.912) & 0.913 (0.910--0.916) & 0.930 (0.929--0.932) \\
Phikon-v2 & 0.909 (0.905--0.912) & 0.916 (0.913--0.918) & 0.929 (0.928--0.931) \\
Shazam & \textbf{0.928 (0.925--0.930)} & \underline{0.931 (0.929--0.934)} & \underline{0.945 (0.944--0.947)} \\
\bottomrule
\end{tabular}
\end{table*}

\begin{table*}[!ht]
\centering
\caption{\textbf{Tile-level classification performance of different foundation models on Chaoyang dataset.} Non-parametric bootstrapping with 1,000 replicates is employed for statistical analysis. The 95\% CI is included in parentheses. Best performing model is \textbf{bolded} and second-best is \underline{underlined}.}
\label{tab:Table18}
\begin{tabular}{lccc}
\toprule
Model & Balanced ACC & Weighted F1 & Top1-ACC \\
\midrule
UNI 2 & 0.767 (0.745--0.787) & 0.767 (0.745--0.784) & 0.820 (0.802--0.833) \\
Virchow2 & \underline{0.821 (0.802--0.840)} & \textbf{0.823 (0.805--0.841)} & \textbf{0.868 (0.853--0.881)} \\
H-optimus-1 & 0.801 (0.780--0.820) & 0.798 (0.780--0.817) & 0.842 (0.826--0.858) \\
Prov-Gigapath & 0.786 (0.765--0.807) & 0.791 (0.771--0.810) & 0.842 (0.825--0.856) \\
Phikon-v2 & 0.765 (0.744--0.785) & 0.776 (0.754--0.795) & 0.834 (0.816--0.849) \\
Shazam & \textbf{0.824 (0.805--0.844)} & \underline{0.820 (0.800--0.838)} & \underline{0.862 (0.846--0.876)} \\
\bottomrule
\end{tabular}
\end{table*}

\begin{table*}[!ht]
\centering
\caption{\textbf{VQA Performance on the PathVQA Dataset.}
The 95\% CI is included in parentheses. The best and second-best performed models are \textbf{bolded} and \underline{underlined.}}
\label{tab:PathVQA_top1}
\begin{tabular}{lc}
\toprule
Model & ACC \\
\midrule
UNI 2      & \underline{0.565 (0.554--0.578)} \\
Virchow2    & 0.559 (0.547--0.571) \\
H-optimus-1   & 0.551 (0.539--0.563) \\
Prov-Gigapath    & 0.562 (0.550--0.574) \\
Phikon-v2  & 0.532 (0.520--0.545) \\
Shazam      & \textbf{0.575 (0.563--0.587)} \\
\bottomrule
\end{tabular}
\end{table*}


\begin{table}[t]
\centering
\small
\caption{\textbf{Effect of teacher ensemble diversity on PCC performance in HEST-Bowel}. The best and second-best performed models are \textbf{bolded} and \underline{underlined.}}
\label{tab:ablation_pcc_hest_bowel}
\begin{tabular}{lc}
\toprule
\textbf{Method} & \textbf{PCC} $\uparrow$ \\
\midrule
\multicolumn{2}{l}{\textbf{Single-teacher models}} \\
UNI-v2 & 0.193 $\pm$ 0.054 \\
Virchow2 & 0.186 $\pm$ 0.040 \\
H-optimus-1 & 0.165 $\pm$ 0.048 \\
Phikon-v2 & 0.156 $\pm$ 0.065 \\
GigaPath & 0.132 $\pm$ 0.100 \\
\midrule
\multicolumn{2}{l}{\textbf{Teacher diversity ablation}} \\
w/o UNI-v2 & 0.213 $\pm$ 0.060 \\
w/o Virchow2 & \underline{0.233 $\pm$ 0.076} \\
w/o UNI-v2, Virchow2 & 0.204 $\pm$ 0.089 \\
w/o UNI-v2, Virchow2, H-optimus-1 & 0.186 $\pm$ 0.100 \\
\midrule
\multicolumn{2}{l}{\textbf{Full teacher ensemble}} \\
Shazam & \textbf{0.242 $\pm$ 0.083} \\
\bottomrule
\end{tabular}
\end{table}

\begin{table*}[!ht]
\centering
\caption{\textbf{Complementarity of multi-level features for WSI-level survival prediction on TCGA-COADREAD.}
The 95\% CI is included in parentheses. The best and second-best performed models are \textbf{bolded} and \underline{underlined.}}
\label{tab:SurvivalAblation}
\begin{tabular}{lc}
\toprule
Model  &  C-index \\
\midrule

\multicolumn{2}{l}{\textbf{Partial feature combinations}} \\
Low only & 0.680 (0.629--0.731) \\
Mid only & 0.689 (0.636--0.743) \\
High only & 0.686 (0.638--0.735) \\
Low + High & \underline{0.702 (0.618--0.785)} \\
Low + Mid & 0.692 (0.638--0.747) \\
Mid + High & 0.699 (0.602--0.796) \\
\midrule
\multicolumn{2}{l}{\textbf{Full multi-level feature ensemble}} \\
Shazam (Low + Mid + High) & \textbf{0.710 (0.655--0.765)} \\
\bottomrule
\end{tabular}
\end{table*}

\begin{table*}[!ht]
\centering
\caption{\textbf{Ablation study evaluating the Mixture of Experts (MoE) module in Shazam on the CRC-100K dataset.} 
Best performance is shown in \textbf{bold}, and the second-best is \underline{underlined}.}
\label{tab:TileAblationPCAM}
\begin{tabular}{lccc}
\toprule
Model & Balanced ACC & Weighted F1 & Top1-ACC \\
\midrule
Shazam  & \textbf{0.960 (0.955--0.966)} & \textbf{0.956 (0.949--0.962)} & \textbf{0.972 (0.968--0.977)} \\
\midrule
\multicolumn{4}{l}{\textbf{MoE Ablation}} \\
w/o MoE & \underline{0.955 (0.950--0.960)} & \underline{0.951 (0.946--0.956)} & \underline{0.970 (0.967--0.974)} \\
\bottomrule
\end{tabular}
\end{table*}

\begin{table*}[!ht]
\centering
\caption{\textbf{Ablation study evaluating the Mixture of Experts (MoE) module in Shazam on the CCRCC dataset}. 
Best performance is shown in \textbf{bold}, and the second-best is \underline{underlined}.}
\label{tab:TileAblationCCRCC}
\begin{tabular}{lccc}
\toprule
Model & Balanced ACC & Weighted F1 & Top1-ACC \\
\midrule
Shazam & \textbf{0.963 (0.953--0.971)} & \textbf{0.975 (0.971--0.978)} & \textbf{0.975 (0.971--0.979)} \\
\midrule
\multicolumn{4}{l}{\textbf{MoE Ablation}} \\
w/o MoE & \textbf{0.963 (0.954--0.971)} & \underline{0.973 (0.969--0.977)} & 0.973 (0.969--0.977) \\
\bottomrule
\end{tabular}
\end{table*}